\documentclass[11pt,letterpaper]{article}

\usepackage[table]{xcolor}

\usepackage[letterpaper,margin=1in]{geometry}
\usepackage[T1]{fontenc}
\usepackage[utf8]{inputenc}
\usepackage{mathptmx}      
\usepackage{helvet}        
\usepackage{graphicx}
\usepackage{xcolor}
\usepackage{titlesec}
\usepackage[most]{tcolorbox}
\usepackage{booktabs}
\usepackage{amsmath,amssymb}
\usepackage[numbers,sort&compress]{natbib}
\usepackage[colorlinks=true]{hyperref}
\usepackage{multirow}
\usepackage{array}
\usepackage{caption}
\usepackage{algorithm}
\usepackage{algorithmic}
\usepackage{threeparttable}
\usepackage{newfloat}
\usepackage{listings}
\usepackage{tabularx}
\usepackage{wrapfig}

\definecolor{brandIndigo}{HTML}{4F46E5}
\definecolor{brandPurple}{HTML}{7C3AED}
\definecolor{brandPink}{HTML}{DB2777}
\definecolor{boxBg}{HTML}{F4F2FD}
\definecolor{boxLine}{HTML}{D9D2F5}

\definecolor{tabBg}{HTML}{E6E2F9}
\definecolor{ablBg}{HTML}{E8E8E8}

\hypersetup{linkcolor=brandPurple, citecolor=brandPurple, urlcolor=brandPurple}

\newcommand{\projecttitle}{\textcolor{brandPurple}{V}ision-\textcolor{brandPurple}{L}anguage-\textcolor{brandPurple}{D}elta-\textcolor{brandPurple}{A}ction: Learning Compact Deltas as Actionable Transition Interfaces for Robotic Manipulation}
\newcommand{\projecturl}{https://baidu-baige.github.io/LoongForge/projects/vlda.html}
\newcommand{\codeurl}{https://github.com/baidu-baige/LoongForge/tree/master/cookbook/loongforge-vlda}
\newcommand{\pageurl}{https://baidu-baige.github.io/LoongForge}
\newcommand{\projectlogofile}{Figures/loongforge-banner.png}  
\newcommand{\projecticonfile}{Figures/hit-icon.png}

\titleformat{\section}{\normalfont\sffamily\large\bfseries}{\thesection}{0.6em}{}
\titleformat{\subsection}{\normalfont\sffamily\normalsize\bfseries}{\thesubsection}{0.6em}{}
\titleformat{\subsubsection}{\normalfont\sffamily\normalsize\bfseries}{\thesubsubsection}{0.6em}{}
\titleformat{\paragraph}[runin]
  {\normalfont\bfseries}
  {}
  {0pt}
  {}

\titlespacing*{\paragraph}
  {0pt}
  {0.5em}
  {0.6em}

\begin{document}
\pagestyle{plain}

\thispagestyle{empty}
\noindent
\includegraphics[height=1.05cm]{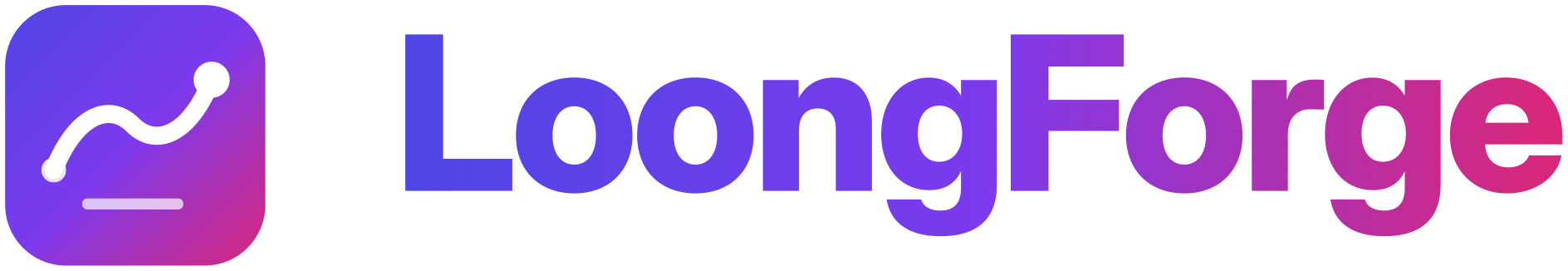}%
\hfill
\includegraphics[height=1.15cm]{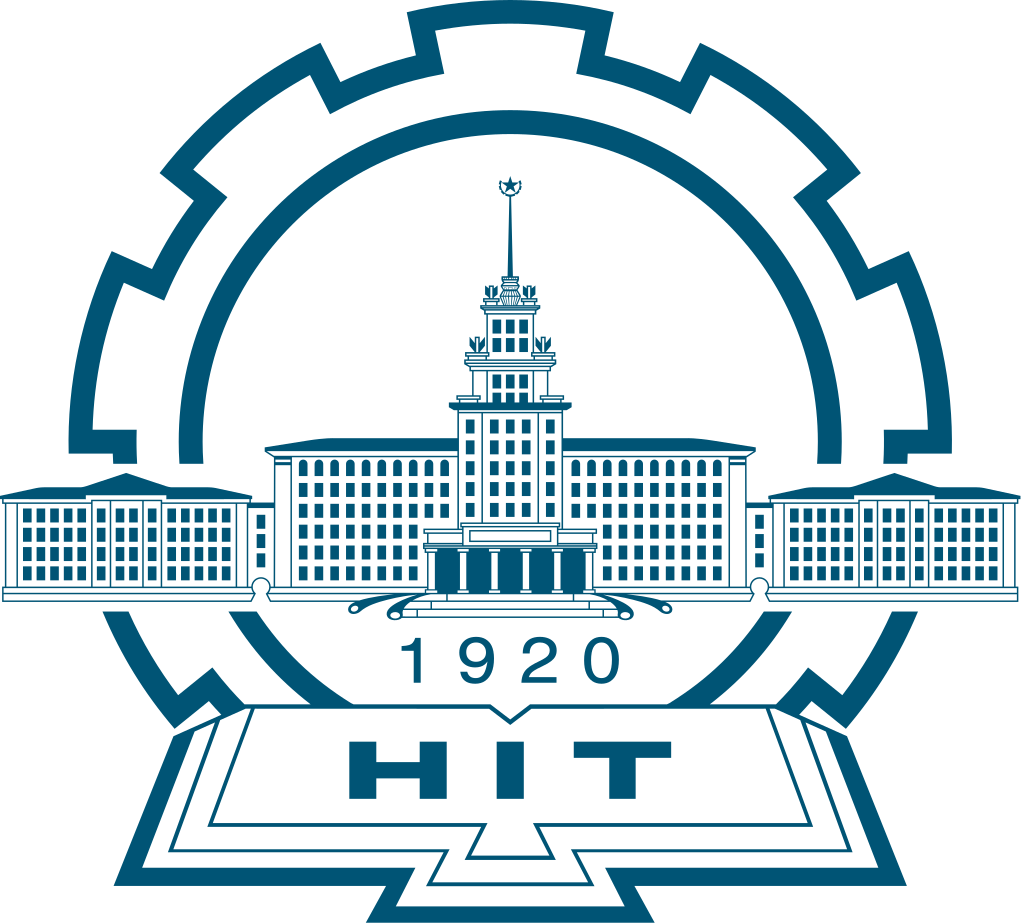}\\[-8pt]
\noindent\textcolor{boxLine}{\rule{\textwidth}{1pt}}

\vspace{1.2em}
\begin{center}
  {\fontsize{22}{26}\selectfont\bfseries \projecttitle\par}
\end{center}

\vspace{0.6em}
\begin{center}
  {\normalsize
   Lingsen Zhang$^{1,2,*}$ \quad Wei Li$^{1,2,*}$ \quad Jie He$^{1,2}$ \quad Henghua Zhang$^{2}$\\[3pt]
   Kai Mo$^{2}$ \quad Shiyong Li$^{2}$ \quad Yanpeng Wang$^{2}$ \quad Rui Shao$^{1,3,\dagger}$\par}
  \vspace{6pt}
  {\small $^{*}$Equal Contribution \quad $^{\dagger}$Corresponding Author\par}
  \vspace{4pt}
  {\small
   $^{1}$Harbin Institute of Technology, Shenzhen \quad
   $^{2}$Baidu, Inc. \quad
   $^{3}$Shenzhen Loop Area Institute \par}
\end{center}

\vspace{0.8em}


\begin{tcolorbox}[colback=boxBg, colframe=boxLine, boxrule=0.6pt,
  arc=3pt, left=12pt, right=12pt, top=10pt, bottom=10pt]
  \small
    Robotic manipulation is not merely observation-to-action generation, but the realization of task-specified state changes in the physical world. Existing vision-language-action (VLA) methods directly learn observation-to-action mappings, imposing a substantial learning burden on policies due to the large gap between semantic understanding and action execution.
    Although world-action models (WAMs) improve control through future prediction, they optimize the full future rather than task-relevant changes, wasting capacity on static appearance or action-irrelevant abstractions.
    To address this issue, we propose \textit{\textbf{Vision-Language-Delta-Action (VLDA)}}, a new paradigm that reformulates full future prediction as compact actionable transition modeling.
    Our core insight is to introduce a \textit{\textbf{compact Delta}} as an intermediate decision variable that captures latent world-state changes, serving as a unified representational interface between vision-language understanding and action generation.
    To learn such a Delta representation, VLDA introduces a unified objective composed of the \textit{\textbf{Local Transition Filter}} (LT-Filter) and \textit{\textbf{Global Transition Anchor}} (GT-Anchor). LT-Filter captures task-relevant manipulation-centric changes, preventing Delta from being biased toward static patch reconstruction. GT-Anchor provides coarse future-level regularization, ensuring that Delta can explain the overall trend of future changes.
    Building on these, \textit{\textbf{Delta-Driven Action Refinement}} (DDAR) further grounds Delta in the action manifold and injects it as a structured seed to guide action denoising.
    We implement VLDA within \textcolor{brandPurple}{LoongForge}, an open-source training framework for large-scale foundation and embodied models, leveraging its modular and distributed infrastructure for unified two-stage optimization.
    Experiments show that VLDA improves long-horizon consistency, robustness, and generalization, validating actionable Delta as a more principled alternative to both direct action prediction and full future-state modeling.

  \vspace{6pt}
  \noindent\textbf{Code:}~\url{\codeurl}
  
  \noindent\textbf{Project:}~\url{\projecturl}
\end{tcolorbox}

\vspace{0.6em}


\section{Introduction}
\label{sec:intro}
\begin{wrapfigure}{r}{0.48\textwidth}
    \centering
    \includegraphics[width=\linewidth]{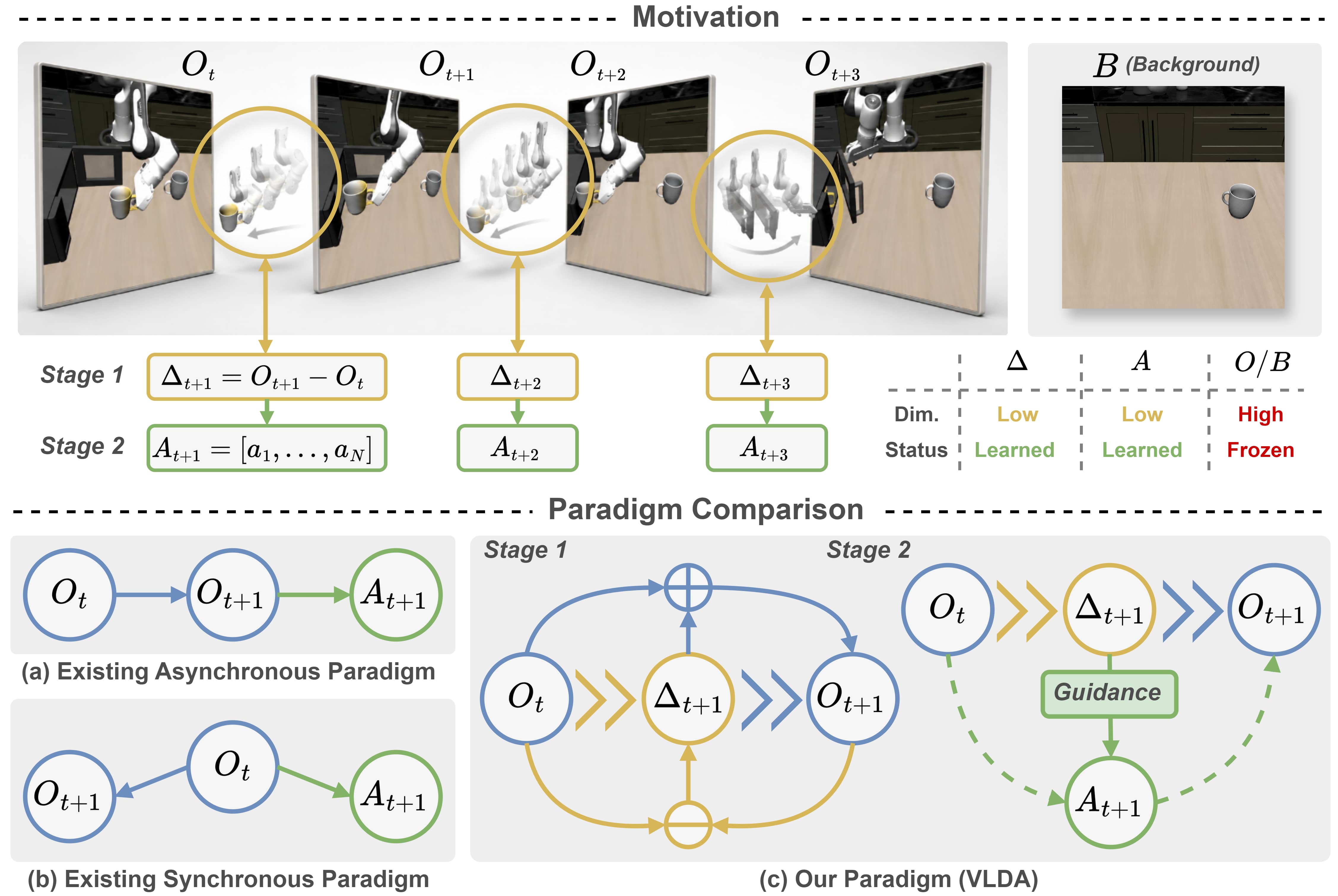}
    \vspace{-8pt}
    \caption{\textbf{Motivation and paradigm comparison of VLDA.}
    Robotic manipulation is governed by sparse task-relevant state changes
    that provide essential cues for action generation, while most visual
    content remains static. Unlike existing asynchronous and synchronous
    paradigms that predict complete future states before or alongside actions,
    VLDA learns a compact \textbf{Delta} from current-to-future transitions
    and uses it as an actionable interface to guide action refinement,
    forming an understanding-transition-execution loop.}
    \label{fig:intro}
    \vspace{-8pt}
\end{wrapfigure}

Similar to human actions, robotic manipulation is a process of controlled physical state transformation~\cite{zeng2021transporter, wang2016actions}. 
As shown in Fig.~\ref{fig:intro} (top), conditioned on visual observations, a robot must not only recognize the current scene and interpret the task-relevant physical context around it~\cite{oft, pi05}, but also determine which part of the scene should change and how such change can be realized through continuous action execution~\cite{souvcek2024genhowto, alayrac2017joint}. 
This requirement fundamentally distinguishes robotic manipulation from generic vision-language understanding, as the critical decision variable is not the complete visual state itself, but the task-relevant transition that drives the current state toward the desired physical outcome.

However, existing VLA and WAM paradigms share a common limitation in lacking a dedicated, action-grounded representation of task-relevant physical transitions.
\textit{\textbf{1)}} Most VLA models still follow an observation-to-action formulation that entangles task-relevant object identification~\cite{openvla, pi0, yuan2026qwen, zheng2025x}, desired physical state-change inference, and executable action-chunk generation within a single policy. 
Without dedicated modeling of task-relevant state changes, the policy must implicitly infer the intended physical transition from action supervision alone, leaving a substantial representational gap between semantic understanding and action execution.

A straightforward way to bridge semantic understanding and action execution is future modeling, as illustrated in Fig.~\ref{fig:intro} (a) and (b). 
\textit{\textbf{2)}} WAMs predict future observations~\cite{cotvla, cen2025worldvla, yuan2026fast}, visual latents~\cite{sun2026vlajepa, lu2024manigaussian}, or action-conditioned dynamics~\cite{vjepa2, jiang2025enerverse} to provide temporal cues beyond the current observation.
However, they typically represent the future as another complete state, causing task-relevant transition cues to remain entangled with static scene content and action-irrelevant variations. Moreover, future prediction is often used only as an auxiliary objective, without providing a structured intermediate variable that can directly guide action generation. 
Robotic manipulation instead requires a compact transition variable that captures task-relevant manipulation-centric changes, remains consistent with the overall direction of future evolution, and is directly grounded in executable actions.

Taken together, these limitations motivate a transition interface that is selective, predictive, and actionable. It should isolate task-relevant manipulation-centric changes, remain consistent with future evolution, and be directly consumable by the action generator. To this end, we propose \textit{\textbf{Vision-Language-Delta-Action (VLDA)}}, a new paradigm that models actionable transitions through a low-dimensional Delta interface rather than predicting the complete future state.

Specifically, VLDA instantiates the transition interface as a compact low-dimensional \textit{\textbf{Delta}} that captures the task-relevant transition implied by the task. To make Delta both selective and predictive, VLDA introduces a unified learning objective with two complementary constraints.
\textit{\textbf{1) Local Transition Filter (LT-Filter)}} distills sparse, manipulation-centric transition evidence from dense current-to-future visual differences, suppressing static scene content and action-irrelevant variations.
\textit{\textbf{2) Global Transition Anchor (GT-Anchor)}} regularizes Delta with a coarse summary of future evolution, preventing overly local transition modeling and preserving the global direction of state change. 
Together, these constraints reconcile local transition selectivity with global predictive consistency, yielding a compact and future-grounded representation of task-relevant physical change.
To make Delta directly actionable, VLDA further introduces \textit{\textbf{3) Delta-Driven Action Refinement (DDAR)}}, which grounds Delta in the action manifold through Delta-action alignment and injects it into the action flow as a structured transition seed.
The action expert then refines this transition-informed seed into executable action chunks conditioned on the predicted transition.
Rather than treating future prediction as auxiliary supervision, VLDA therefore establishes a closed \textit{\textbf{understanding-transition-execution loop}}, where vision-language representations interpret the task context, Delta specifies the required state transition, and DDAR realizes that transition through continuous action generation.

We evaluate VLDA on standard robotic manipulation benchmarks and challenging long-horizon tasks. Experimental results show that VLDA improves long-horizon consistency, robustness under distribution shifts, and generalization to complex manipulation scenarios. These results demonstrate that modeling actionable Delta provides a more effective alternative to both direct observation-to-action learning and full future-state prediction.
Our contributions are summarized as follows:
\begin{itemize}
\setlength{\itemsep}{0pt}
\setlength{\parskip}{0pt}
\setlength{\parsep}{0pt}
    \item We introduce \textit{\textbf{Vision-Language-Delta-Action (VLDA)}}, a new robotic manipulation paradigm that replaces complete future-state prediction with a compact \textit{\textbf{Delta}} serving as an actionable transition interface.
    \item We design a unified Delta learning objective with \textit{\textbf{Local Transition Filter}} and \textit{\textbf{Global Transition Anchor}}, enabling Delta to capture sparse task-relevant changes while remaining consistent with future transition trends.
    \item We develop \textit{\textbf{Delta-Driven Action Refinement}}, which aligns Delta with the action manifold and uses it as a structured seed for action denoising, forming an explicit understanding-transition-execution loop.
\end{itemize}

\section{Related Work}
\label{sec:related}
\paragraph{Observation-to-Action Learning in VLA Models.}
Existing VLA models primarily employ action-token prediction, implemented either autoregressively or in parallel~\cite{Rt-2, openvla, oft, hu2026ar, shao2025large, li2026consisvla}, and continuous action experts based on diffusion or flow matching~\cite{diff, pi0, pi05, wen2025dexvla, li2026unimpa, hu2026abstraction, li2026global}.
Despite different action decoders, both paradigms directly supervise actions from observations, requiring a single policy to jointly resolve task grounding, intended physical change, and motor realization. This creates a substantial gap between high-level semantic observations and low-level executable actions. VLDA bridges this gap with a compact {Delta} that explicitly mediates what should change and how that change is executed.

\paragraph{Predictive World-Action Modeling.}
Recent WAMs couple future modeling with robot control through cascaded future-to-action generation~\cite{tian2024predictive, li2026causal, zhu2026h}, joint world-action prediction~\cite{guo2024prediction, cen2025worldvla, yuan2026fast, zhu2026learning}, and predictive latent dynamics~\cite{hu2024video, huang2025ladi, miao2026jepavla}.
In these formulations, future prediction typically serves as an upstream context, a jointly decoded target, or a regularizer for action learning, with its contribution to control determined by architecture-specific fusion mechanisms.
VLDA instead reparameterizes the prediction-control connection through a compact Delta, whose control role is explicitly defined by transition learning and Delta-driven action refinement.

\paragraph{Latent Action Representation Learning.}
Discrete approaches tokenize visually inferred motion patterns for scalable VLA pretraining~\cite{ye2024latent, univla, chen2025villa}, while continuous formulations learn smooth latent control variables that can be mapped to embodiment-specific actions with limited supervision~\cite{liang2025clam, tharwat2025latent}. CLAP~\cite{zhang2026clap} and LARA~\cite{liu2026lara} further improve physical grounding by aligning visually inferred representations with robot actions or proprioceptive dynamics. However, these representations mainly serve as pseudo-action labels, pretraining targets, or shared control spaces. They do not explicitly isolate task-relevant physical changes under complementary local and global constraints, ground such changes in the action manifold, and use them to directly guide action refinement.

\begin{figure*}[t]
\centering
\includegraphics[width=1.0\textwidth]{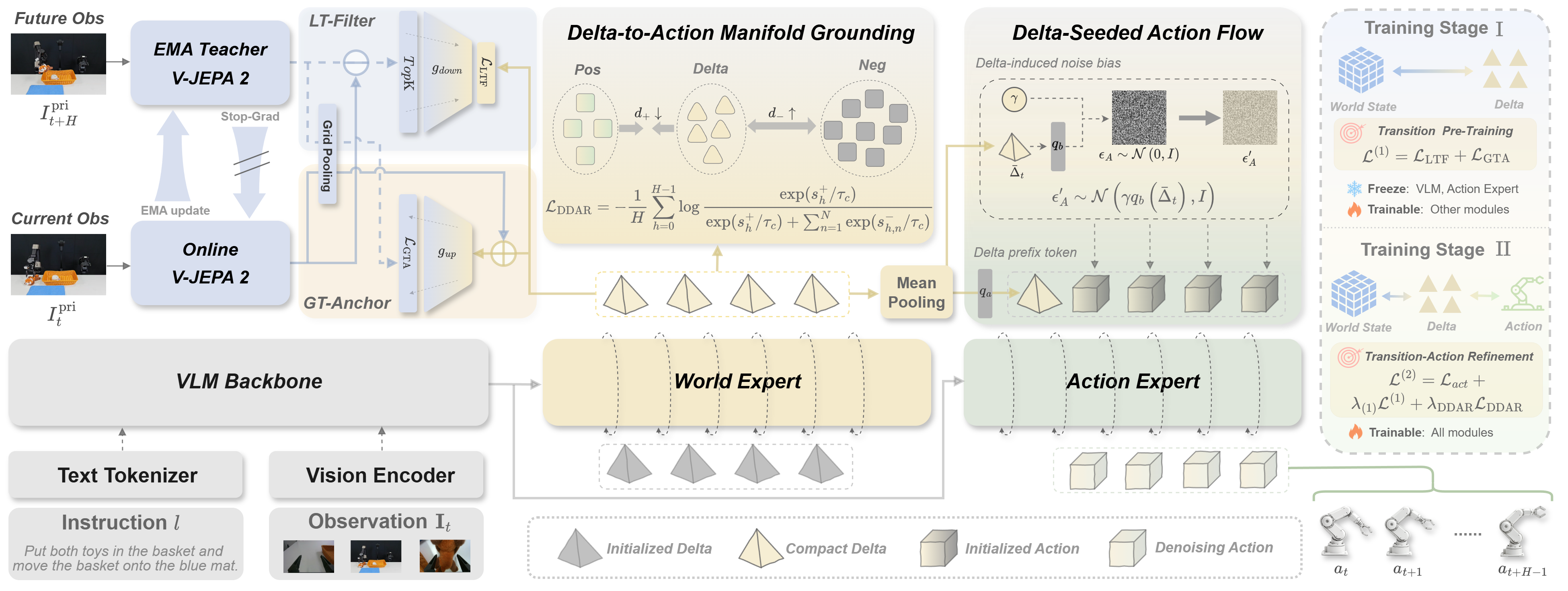} 
\caption{\textbf{Overview of VLDA.}
Given the observation and instruction, the World Expert predicts an actionable Delta under transition supervision. In training stage I, LT-Filter and GT-Anchor learn selective and consistent transition representations. In stage II, Delta-Driven Action Refinement grounds Delta in the action manifold through Delta-to-Action Manifold Grounding and injects it into action denoising through Delta-Seeded Action Flow, enabling the Action Expert to generate action chunks.}
\label{fig:main}
\vspace{-10pt}
\end{figure*}

\section{Preliminary \& Problem Definition}
\label{sec:preliminary}

\paragraph{Problem Definition.}
We consider robotic manipulation at step $t$, where the robot receives an instruction $l$ and observes $s_t=(\mathbf{I}_t,p_t)$. Here, $\mathbf{I}_t=\{I_t^c\}_{c=1}^{C}$ denotes multi-view RGB observations and $p_t$ denotes proprioception. The policy predicts an action chunk $\mathbf{A}_t=[a_t,\ldots,a_{t+H-1}] \in\mathbb{R}^{H\times d_a}$, where $H$ and $d_a$ denote the action horizon and dimension, respectively. During training, the corresponding future primary-camera image $I_{t+H}^{\mathrm{pri}}$ is available, while inference only relies on $(s_t,l)$. Our objective is to exploit such training-time state evolution to improve $p(\mathbf{A}_t\mid s_t,l)$ without requiring future observations at deployment.

\paragraph{Conditional Flow Matching.}
Given a clean target $\mathbf{x}_0$ and source $\mathbf{z}$ under condition $c$, both experts use the same linear conditional path and corresponding target velocity:
\begin{equation}
    \mathbf{x}_\tau
    =
    (1-\tau)\mathbf{x}_0+\tau\mathbf{z},
    \qquad
    \mathbf{u}_\tau^\star
    =
    \mathbf{z}-\mathbf{x}_0.
\end{equation}
The Action Expert uses the biased source $\mathbf{z}_A=\epsilon_A+\gamma\mathbf{b}(c)$ and directly predicts $\hat{\mathbf{u}}_\tau$. The World Expert uses
$\mathbf{z}_\Delta=\epsilon_\Delta$ and $x$-prediction, estimating
$\hat{\mathbf{x}}_0=f_\vartheta(\mathbf{x}_\tau,\tau,c)$ and recovering velocity as
\begin{equation}
    \hat{\mathbf{u}}_\tau
    =
    \frac{\mathbf{x}_\tau-\hat{\mathbf{x}}_0}{\tau},
\end{equation}
where $\epsilon_A,\epsilon_\Delta \sim\mathcal{N}(0,I)$ denote the noise variables used by the Action and World Experts, respectively.

\paragraph{Exponential Moving Average.}
EMA maintains a slowly evolving model copy to provide stable supervision targets. Let $\psi_r$ and $\bar{\psi}_r$ denote the online and EMA weights at iteration $r$, respectively. The EMA weights are recursively updated at every training iteration as
\begin{equation}
    \bar{\psi}_r
    =
    m\bar{\psi}_{r-1}
    +
    (1-m)\psi_r, ~~ m \in [0,1),
\end{equation}
where $m$ is the EMA momentum governing the temporal smoothing of teacher weights. The EMA model is not optimized by gradients and evolves smoothly during training.

\section{Vision-Language-Delta-Action}
\subsection{Framework: Actionable Delta Modeling}
Given the current state $s_t$ and language instruction $l$, conventional VLA policies directly predict an action chunk $\mathbf{A}_t$ as $\mathbf{A}_t \sim \pi(\cdot \mid s_t,l)$. However, such a direct observation-to-action mapping requires a single policy to bridge high-level semantic understanding and low-level action execution \textbf{within a unified end-to-end decision-making process}, making long-horizon manipulation difficult to optimize.

To alleviate this burden, VLDA reformulates action generation as an \textit{understanding--transition--execution} process. A VLM backbone first encodes the state-instruction pair into a multimodal context $\mathbf{h}_t = \Phi_{\mathrm{VLM}}(s_t,l)$. VLDA then introduces a compact actionable Delta $\Delta_t$ as an intermediate transition variable that captures task-specified physical state changes. The action generation process is factorized as:
\begin{equation}
\label{equ:probabilistic}
    p(\mathbf{A}_t \mid s_t,l)
    =
    \int p_\theta(\Delta_t \mid \mathbf{h}_t)
    p_\phi(\mathbf{A}_t \mid \mathbf{h}_t, \Delta_t)
    \,\mathrm{d}\Delta_t .
\end{equation}
Here, $p_\theta$ denotes the World Expert for transition prediction, while $p_\phi$ denotes the Action Expert for action generation conditioned jointly on both the inferred Delta and multimodal context.

Through this factorization, Delta serves as a compact transition interface between vision-language understanding and action generation, replacing redundant full future-state modeling with focused task-relevant state-change modeling for downstream action prediction in diverse robotic manipulation scenarios.

\subsection{Actionable Transition Generation}
\label{sec:world_expert}
As established in Equation~\eqref{equ:probabilistic}, the World Expert $p_\theta$ predicts the actionable Delta $\Delta_t$ from the multimodal context $\mathbf{h}_t$. We model $p_\theta$ as a continuous-time flow-matching model in the Delta space, which generates a clean transition estimate $\hat{\Delta}_t$ while preserving the compact structure of the learned transition representation.

Inspired by the manifold perspective revisited in JiT~\cite{li2025jit}, where natural data lies on a low-dimensional structured manifold while noise spans an unstructured high-dimensional space, we note that physical state changes in robotic manipulation are similarly constrained by scene geometry, kinematics, and object interactions rather than being arbitrarily distributed in the latent space. Thus, we adopt $x$-prediction parameterization for transition generation. Given the noisy Delta state $\Delta_\tau$ at flow time $\tau$, the model predicts
\begin{equation}
    \hat{\Delta}_t = f_\theta^\Delta(\Delta_\tau, \tau, \mathbf{h}_t).
\end{equation}
The $\Delta_\tau$ is introduced in Section~\ref{sec:LT_Filter}, with its detailed construction deferred to the transition-learning objective below. Following the $x$-prediction form, the corresponding velocity is analytically recovered as
\begin{equation}
    \hat{\mathbf{u}}_\tau^\Delta = \frac{\Delta_\tau - \hat{\Delta}_t}{\tau}.
\end{equation}
The predicted $\hat{\Delta}_t$ serves as a compact transition prior for action generation. To ensure that this prior captures task-relevant physical changes rather than collapsing into trivial representations, we introduce two complementary objectives: Local Transition Filter and Global Transition Anchor, which respectively emphasize selective local changes and coarse global consistency within the learned transition representation.

\subsection{Local Transition Filter}
\label{sec:LT_Filter}
Learning the transition prior for the World Expert necessitates a supervision target that isolates actionable physical changes from low-level visual redundancies for stable task-relevant transition learning. Instead of relying on raw pixels, we extract dense semantic features from the current and future observations:
\begin{equation}
    \mathbf{V}_t
    =
    E_{\mathrm{on}}(I_t^{\mathrm{pri}}),
    ~~
    \bar{\mathbf{V}}_{t+H}
    =
    E_{\mathrm{ema}}(I_{t+H}^{\mathrm{pri}})
    \in \mathbb{R}^{P\times d_v}.
\end{equation}
where $P$ denotes the number of spatial tokens. Specifically, $E_{\mathrm{on}}$ and $E_{\mathrm{ema}}$ denote the two encoders, with comprehensive elaborations provided in Section~\ref{sec:training_objective}. The dense transition is computed as $\mathbf{D}_t = \bar{\mathbf{V}}_{t+H} - \mathbf{V}_t.$

Rather than reconstructing the entire future feature map, which contains substantial static redundancy, LT-Filter isolates the task-relevant dynamics by selecting the top-$K$ regions with the largest transition magnitudes, thereby focusing supervision on regions exhibiting the strongest transition evidence:
\begin{equation}
\begin{aligned}
    \mathcal{K}_t 
    &= \operatorname{TopK}_{K}
    \left(\{\|\mathbf{D}_{t,i}\|_2\}_{i=1}^{P}\right), \\
    \Delta_t^\star 
    &= g_{\downarrow}\left(\mathbf{D}_{t,\mathcal{K}_t}\right)
    \in \mathbb{R}^{K \times d_\Delta}.
\end{aligned}
\end{equation}
Here, $g_{\downarrow}$ projects the selected transition tokens into a compact Delta space, with $K \ll P$ and $d_\Delta \ll d_v$, which keeps the representation compact while retaining task-relevant transition evidence. We refer to $\Delta_t^\star$ as the Delta target, which filters out static appearance information while retaining local changes that are most likely to be actionable. Given $\Delta_t^\star$ and Gaussian noise $\epsilon_\Delta\sim\mathcal{N}(0,I)$, the Delta flow path is constructed as
\begin{equation}
    \Delta_\tau = \tau\epsilon_\Delta + (1-\tau)\Delta_t^\star,
    ~~
    \mathbf{u}_\tau^\Delta = \epsilon_\Delta - \Delta_t^\star.
\end{equation}

Since the tokens indexed by $\mathcal{K}_t$ form an unordered set, standard point-wise regression is permutation-sensitive. We therefore use an entropy-regularized transport plan to construct a balanced soft correspondence. For two token sets $\mathbf{X},\mathbf{Y}\in\mathbb{R}^{K\times d}$, we define the pairwise cost as \begin{equation} 
    C_{ij} = \frac{\|\mathbf{x}_i-\mathbf{y}_j\|_2^2}{d}. 
\end{equation}
To resolve the ambiguity caused by the unordered token structure, we formulate transition matching as a soft set-to-set assignment problem rather than enforcing fixed index-wise correspondence. The pairwise costs therefore characterize the compatibility between predicted and target transition tokens, while a balanced transport plan determines their soft correspondence in a permutation-invariant manner. The transport plan and set-matching loss are then jointly computed as follows:
\begin{equation} 
\small 
\begin{aligned} 
    \mathbf{T}_\eta^\star 
    &= 
    \operatorname{Sinkhorn}_\eta(\bar{\mathbf{C}}),\\ 
    \mathcal{D}_{\mathrm{set}}(\mathbf{X},\mathbf{Y};d) 
    &= 
    \left\langle \operatorname{sg}(\mathbf{T}_\eta^\star), \mathbf{C} \right\rangle, 
\end{aligned} 
\label{equ:Dset} 
\end{equation} 
where $\bar{\mathbf{C}}$ is the sample-wise normalized cost, $\operatorname{sg}$ denotes stop-gradient, and $\eta$ controls the softness of the balanced transport plan. The Local Transition Filter objective is then 
\begin{equation} 
\label{eq:LOSS_LTF} 
    \mathcal{L}_{\mathrm{LTF}} = \mathbb{E}_{\tau,\epsilon_\Delta} \left[ \mathcal{D}_{\mathrm{set}} \left( \hat{\mathbf{u}}_\tau^\Delta, \mathbf{u}_\tau^\Delta; d_\Delta \right) \right]. 
\end{equation} 
Further details are provided in \textbf{Appendix~D.2}, including implementation and optimization specifics.

\subsection{Global Transition Anchor}
\label{sec:GT_Anchor}
While LT-Filter emphasizes fine-grained local regions with prominent state changes, relying only on the selected top-$K$ regions risks losing the coarse global context of the overall transition. To preserve global consistency, we introduce the Global Transition Anchor (GT-Anchor) as a coarse transition-level regularization, thereby complementing local selectivity with a scene-level constraint on overall transition direction.

Specifically, we aggregate the dense visual features $\mathbf{V}_t$ and $\bar{\mathbf{V}}_{t+H}$ into $K$ coarse tokens via adaptive pooling, and construct a global transition anchor that summarizes the overall state evolution direction:
\begin{equation}
    \mathbf{G}_t =
    \operatorname{Pool}_K(\bar{\mathbf{V}}_{t+H})
    -
    \operatorname{Pool}_K(\mathbf{V}_t)
    \in \mathbb{R}^{K \times d_v}.
\end{equation}
The resulting $\mathbf{G}_t$ provides a coarse reference for the transition direction by summarizing large-scale changes across the pooled current and future representations.
Meanwhile, the predicted compact Delta $\hat{\Delta}_t$ is projected back to the high-dimensional visual feature space using an up-projection module $g_{\uparrow}$:
\begin{equation}
    \hat{\mathbf{D}}_t=
    g_{\uparrow}(\hat{\Delta}_t)
    \in \mathbb{R}^{K \times d_v}.
\end{equation}
The GT-Anchor objective matches the projected transition tokens with the global transition anchor using the set distance defined in Equation~\eqref{equ:Dset} to enforce global consistency across transition representations:
\begin{equation}
    \mathcal{L}_{\mathrm{GTA}}
    =
    \mathbb{E}_{\tau,\epsilon_\Delta}
    \left[
    \mathcal{D}_{\mathrm{set}}
    \left(
    \hat{\mathbf{D}}_t,
    \mathbf{G}_t;
    d_v
    \right)
    \right].
\end{equation}

Importantly, GT-Anchor does not require reconstructing the full future state. Instead, it anchors the compact Delta to the coarse direction of future state transitions, which complements LT-Filter by providing a better balance between expressive capacity and structured transition regularization.

\subsection{Delta-Driven Action Refinement}
\label{sec:DDAR}
\paragraph{Delta-Seeded Action Flow (DSAF).} 
Given the predicted $\hat{\Delta}_t$, DDAR guides action denoising
through two complementary branches: a Delta-induced noise bias and
a Delta prefix token. We first aggregate the Delta tokens into a global transition context and map it to an action-space bias:
\begin{equation}
    \bar{\Delta}_t = \frac{1}{K}\sum_{k=1}^{K}\hat{\Delta}_{t,k},
    \quad
    \mathbf{b}_t = q_b(\bar{\Delta}_t) \in \mathbb{R}^{H\times d_a}.
\end{equation}

Instead of standard Gaussian noise, we shift the initial action noise $\epsilon_A\sim\mathcal{N}(0,I)$ with this Delta-induced bias and construct the action flow path, thereby introducing transition-aware structure at initialization:
\begin{equation}
    \epsilon_A' = \epsilon_A + \gamma\mathbf{b}_t,
    \quad
    \mathbf{A}_\tau = \tau\epsilon_A' + (1-\tau)\mathbf{A}_t,
\end{equation}
where $\gamma$ is a learnable scalar controlling the bias strength. This path yields the target action velocity $\mathbf{u}_\tau^A=\epsilon_A'-\mathbf{A}_t$. In the second branch, $\bar{\Delta}_t$ is projected into a Delta prefix token and prepended to the action tokens during denoising. The Action Expert $f_\phi^A$ then predicts the action velocity from the multimodal context, the action flow state, and the prepended Delta context token under this conditioned denoising formulation:
\begin{equation}
    \hat{\mathbf{u}}_\tau^A
    =
    f_\phi^A(\mathbf{A}_\tau,\tau,\mathbf{h}_t,\bar{\Delta}_t).
\end{equation}
The action generation objective is defined as an MSE loss:
\begin{equation}
    \mathcal{L}_{\mathrm{act}}
    =
    \mathbb{E}_{\tau,\epsilon_A}
    \left[
    \|\hat{\mathbf{u}}_\tau^A-\mathbf{u}_\tau^A\|_2^2
    \right].
\end{equation}

In this way, Delta is not only predicted as an auxiliary transition variable, but also injected into the action flow as a physically grounded prior during downstream action generation and iterative denoising.

\begin{table*}[t]
\caption{Results on LIBERO-Plus~\cite{fei2025libero}. We report generalization performance across seven perturbation dimensions and four task suites. VLDA is evaluated in a \textbf{zero-shot} setting with 10,030 rollouts.}
\centering
\footnotesize
\setlength{\tabcolsep}{4pt}
\label{tab:libero_plus_main}
\resizebox{\textwidth}{!}{
    \begin{tabular}{l|ccccccc|cccc|c}
        \toprule
        \multirow{2}{*}{\textbf{Method}} 
        & \multicolumn{7}{c|}{\textbf{Perturbation Dimensions}}
        & \multicolumn{4}{c|}{\textbf{Task Suites}} & \multirow{2}{*}{\textbf{Avg.}} \\
        \cmidrule(lr){2-8} \cmidrule(lr){9-12}
        & Camera & Robot & Language & Light & Background & Noise & Layout
        & Spatial & Object & Goal & Long &  \\
        \midrule
        OpenVLA~\cite{openvla} 
        & 0.8 & 3.5 & 23.0 & 8.1 & 34.8 & 15.2 & 28.5 
        & 19.4 & 14.0 & 15.1 & 14.3 & 15.6 \\
        
        OpenVLA-OFT~\cite{oft}
        & 56.4 & 31.9 & {79.5} & 88.7 & {93.3} & 75.8 & 74.2 
        & 84.0 & 66.5 & 63.0 & 66.4 & 69.6 \\
        
        UniVLA~\cite{univla}
        & 1.8 & 46.2 & 69.6 & 69.0 & 81.0 & 21.2 & 31.9 
        & 55.5 & 36.7 & 40.7 & 39.9 & 43.9 \\
        
        WorldVLA~\cite{cen2025worldvla} 
        & 0.1 & 27.9 & 41.6 & 43.7 & 17.1 & 10.9 & 38.0 
        & 32.5 & 28.6 & 31.8 & 8.2 & 25.0 \\
        
        
        DreamVLA~\cite{zhang2026dreamvla}
        & {65.0} & 40.8 & 63.5 & 85.7 & 82.6 & {84.9} & 74.0 
        & 79.7 & 79.0 & 61.7 & 59.8 & 69.9 \\
        
        
        GuidedVLA~\cite{jia2026guidedvla}
        & \textbf{73.7} & 51.4 & 62.6 & {94.6} & 89.0 & {85.2} & 79.9 
        & 84.0 & 80.9 & {70.8} & 66.2 & {75.4} \\

        $\pi_0$~\cite{pi0}
        & 13.8 & 6.0 & 58.8 & 85.0 & 81.4 & 79.0 & 68.9 
        & 60.7 & 61.4 & 44.9 & 48.4 & 53.6 \\

        
        $\pi_{0.5}$~\cite{pi05}
        & 59.7 & {65.5} & 75.3 & 87.0 & 82.4 & 72.1 & {80.3} 
        & 79.9 & {87.8} & 69.0 & 64.9 & 73.6 \\

        Spatial Forcing~\cite{li2025spatial}
        & 20.1 & 13.4 & 40.9 & 29.1 & 33.4 & 25.7 & 39.3 
        & 52.9 & 31.0 & 28.2 & 5.4 & 29.1 \\
        
        VLA-Adapter~\cite{wang2025vlaadapter}
        & 36.2 & 37.9 & 74.6 & 70.6 & 76.1 & 58.0 & 69.7 
        & 85.0 & 46.3 & 56.0 & 50.4 & 59.1 \\

        \midrule
        \rowcolor{tabBg}
        \textbf{VLDA} 
        & 72.4
        & \textbf{75.3} 
        & \textbf{84.0} 
        & \textbf{96.0} 
        & \textbf{95.5} 
        & \textbf{89.4} 
        & \textbf{85.7} 
        & \textbf{89.6} 
        & \textbf{89.8}
        & \textbf{80.1} 
        & \textbf{78.3} 
        & \textbf{84.5} \\

        \bottomrule
    \end{tabular}
}
\vspace{-5pt}
\end{table*}

\paragraph{Delta-to-Action Manifold Grounding (DAMG).} 
To ensure that Delta captures action-relevant transitions rather than purely visual changes, we align Delta tokens with action tokens in a shared contrastive space to encourage consistent cross-modal action grounding. We treat the $H$ ground-truth actions $\{a_{t+h}\}_{h=0}^{H-1}$ in the action chunk as positive samples, and randomly sample $N$ actions from non-overlapping temporal regions of the same trajectory as negative samples. The Delta and action tokens are projected as:
\begin{equation}
\small
    \mathbf{z}_{k}^{\Delta}
    =
    \operatorname{Norm}(q_\Delta(\hat{\Delta}_{t,k})),
    ~~
    \mathbf{z}_{h}^{\mathrm{act}}
    =
    \operatorname{Norm}(q_{\mathrm{act}}(a_{t+h})).
\end{equation}
For each positive action token $a_{t+h}$, we compute its similarity to all Delta tokens using soft token pooling:
\begin{equation}
\small
    s_h^+ = \tau_\Delta \log \sum_{k=1}^{K}
    \exp \left(
    \frac{
    \langle \mathbf{z}_{k}^{\Delta},
    \mathbf{z}_{h}^{\mathrm{act}} \rangle
    }{\tau_\Delta}
    \right).
\end{equation}
where $\tau_\Delta$ is the pooling temperature controlling soft token aggregation. For the negative actions, we compute their similarities $s_n^-$ to the Delta set in the same way and optimize an InfoNCE objective:
\begin{equation}
\small
    \mathcal{L}_{\mathrm{DDAR}} = -\frac{1}{H} \sum_{h=0}^{H-1} \log \frac{\exp(s_h^+/\tau_c)}{\exp(s_h^+/\tau_c) + \sum_{n=1}^{N}\exp(s_n^-/\tau_c)},
\end{equation}
where $\tau_c$ is the contrastive temperature controlling positive-negative separation. This alignment grounds Delta in the action manifold, encouraging it to be not only visually predictive but also action-executable.

\section{Training Strategy}
\label{sec:training_objective}

\subsection{Training Objective}

Following V-JEPA 2~\cite{vjepa2}, VLDA employs an online encoder $E_{\mathrm{on}}$ for current observations and an EMA teacher $E_{\mathrm{ema}}$ for future observations. Specifically, $E_{\mathrm{on}}$ is optimized by gradients, whereas $E_{\mathrm{ema}}$ is updated solely through exponential moving average (see \textbf{Appendix D.1}). This asymmetric design prevents the Delta target from being constructed between two rapidly changing feature spaces, thereby providing stable supervision for future-transition modeling.

VLDA is optimized in two stages to progressively establish transition modeling and action generation. In the first stage (transition pre-training), we learn the transition interface with the Local Transition Filter and Global Transition Anchor objectives, while keeping the action branch frozen:
\begin{equation}
    \mathcal{L}^{(1)}
    =
    \mathcal{L}_{\mathrm{LTF}}
    +
    \mathcal{L}_{\mathrm{GTA}}.
\end{equation}
In the second stage, we jointly optimize transition modeling, action generation, and Delta--action alignment:
\begin{equation}
    \mathcal{L}^{(2)}
    =
    \mathcal{L}_{\mathrm{act}}
    +
    \lambda_{(1)}\mathcal{L}^{(1)}
    +
    \lambda_{\mathrm{DDAR}}\mathcal{L}_{\mathrm{DDAR}}.
\end{equation}



\subsection{Training Framework}

We implement and train VLDA within LoongForge\footnote{\url{https://baidu-baige.github.io/LoongForge/}}~\cite{LoongForge2026}, an open-source training framework for large-scale foundation and embodied models. Its PyTorch-native embodied-model subsystem provides modular model, data, and trainer abstractions, supporting distributed training with DDP, ZeRO-1, FSDP, and HSDP. Together with flexible component-wise parameter freezing, checkpoint management, and experiment logging, this infrastructure facilitates the staged optimization of VLDA's VLM backbone, World Expert, Action Expert, and transition modules. Building upon these capabilities, we implement VLDA-specific Delta target construction, LT-Filter and GT-Anchor objectives, Delta--action alignment, and Delta-driven action refinement, integrating transition learning and action generation within a unified training pipeline.

During inference, VLDA first samples $\hat{\Delta}_t$ with the World Expert and then uses it as a structured transition prior to guide action denoising via the Action Expert. This inference procedure preserves the understanding--transition--execution decomposition, allowing the predicted Delta to mediate between multimodal task understanding and executable action generation without requiring access to future observations.

\begin{figure*}[t]
\centering
\includegraphics[width=1.0\textwidth]{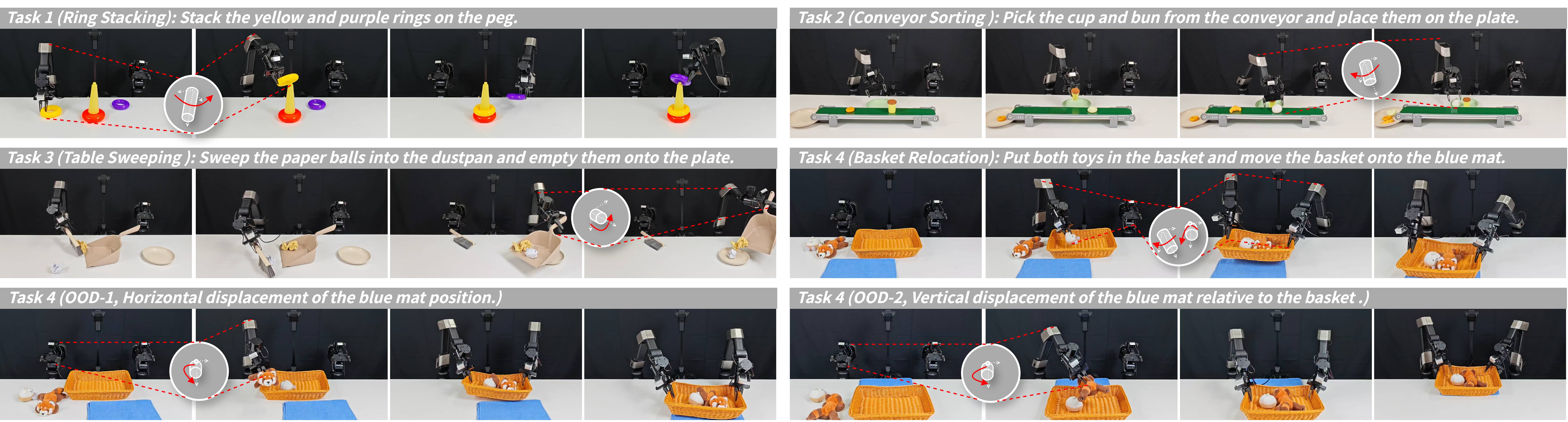} 
\vspace{-15pt}
\caption{Representative executions on Ring Stacking, Conveyor Sorting, Table Sweeping, and Basket Relocation, including two OOD spatial configurations. Frames proceed from left to right, while {red dashed lines} and {circular zoom-ins} highlight the correspondence between task-relevant current-to-future transitions and actions. Complete trajectories in \textbf{Appendix~C.1}.}
\label{fig:vis1}
\vspace{-5pt}
\end{figure*}

\begin{table*}[ht]
\centering
\caption{Results on selected \textbf{hard} tasks from RoboTwin 2.0~\cite{chen2025robotwin}, chosen to cover variations in layout, illumination, appearance, and other factors. VLDA is trained exclusively on the clean setting and evaluated on the randomized setting, constituting a \textbf{zero-shot} evaluation. VLDA achieves the best average performance, and the best result for each task is highlighted in \textbf{bold}. Complete results across all 50 tasks are reported in \textbf{Appendix~B.1} for reference.}

\label{tab:robotwin_main}
\resizebox{\textwidth}{!}{%
\setlength{\tabcolsep}{3pt}
\begin{tabular}{l|*{11}{>{\centering\arraybackslash}m{1.55cm}}}
\toprule
\textbf{Method}
& \shortstack{\textbf{Ranking}\\\textbf{RGB}}
& \shortstack{\textbf{Ranking}\\\textbf{Size}}
& \shortstack{\textbf{Grab}\\\textbf{Roller}}
& \shortstack{\textbf{Open}\\\textbf{Laptop}}
& \shortstack{\textbf{Open}\\\textbf{Microwave}}
& \shortstack{\textbf{Place Dual}\\\textbf{Shoes}}
& \shortstack{\textbf{Put Bottles}\\\textbf{Dustbin}}
& \shortstack{\textbf{Shake}\\\textbf{Bottle}}
& \shortstack{\textbf{Stacking}\\\textbf{Three}}
& \shortstack{\textbf{Stacking}\\\textbf{Two}}
& \textbf{Avg.} \\
\midrule

ACT~\cite{zhao2023act}
& 0\% & 0\% & 25\% & 0\% & 0\% & 0\% & 1\% & 10\% & 0\% & 0\%
& 3.6\% \\

RDT~\cite{liu2024rdt}
& 0\% & 0\% & 43\% & 32\% & 20\% & 4\% & 4\% & 45\% & 17\% & 30\%
& 19.5\% \\

DP~\cite{diff}
& 0\% & 0\% & 0\% & 0\% & 0\% & 0\% & 0\% & 8\% & 0\% & 0\%
& 0.8\% \\

DP3~\cite{Ze2024DP3}
& 0\% & 0\% & 2\% & 7\% & 22\% & 0\% & 21\% & 19\% & 5\% & 6\%
& 8.2\% \\

TwinVLA~\cite{im2025twinvla}
& 0\% & 0\% & 22\% & 17\% & 1\% & 3\% & 4\% & 58\% & 15\% & 11\%
& 13.1\% \\

$\boldsymbol{\pi_0}$~\cite{pi0}
& 5\% & 1\% & 80\% & 46\% & 50\% & 0\% & 13\% & 60\% & 24\% & 41\%
& 32.0\% \\

$\boldsymbol{\pi_{0.5}}$~\cite{pi05}
& 35\% & 14\% & 89\% & 35\% & 37\% & 7\% & 9\%
& \textbf{100\%} & 35\% & 66\%
& 42.7\% \\

HALO~\cite{shou2026halo}
& 7\% & 2\% & 57\% & 37\% & 24\% & 3\% & 13\% & 73\% & 25\% & 49\%
& 29.0\% \\

BagelVLA~\cite{hu2026bagelvla}
& 4\% & 2\% & 41\% & 37\% & 0\% & 0\% & 10\% & 74\% & 13\% & 52\%
& 23.3\% \\

\midrule
\rowcolor{tabBg}

\textbf{VLDA}
& \textbf{53\%} & \textbf{16\%} & \textbf{96\%} & \textbf{60\%}
& \textbf{64\%} & \textbf{34\%} & \textbf{55\%} & \textbf{100\%}
& \textbf{65\%} & \textbf{86\%} & \textbf{62.9\%} \\
\bottomrule
\end{tabular}%
}
\vspace{-5pt}
\end{table*}

\begin{table*}[th]
    \centering
    \caption{
    Real-world results on the \textbf{GALAXEA R1 Lite platform}~\cite{galaxea}. ``$\dagger$'' denotes reproduced results under VLDA settings. Entries report progressive-stage success over 25 trials; \textbf{Avg.} averages final-checkpoint success across tasks. Checkpoint definitions and \textbf{AgileX Cobot Magic}~\cite{aloha} results are in \textbf{Appendix~B.1}.
    }
    \label{tab:real_world_results}
    \scriptsize
    \setlength{\tabcolsep}{1.5pt}
    \resizebox{\textwidth}{!}{
    \begin{tabular}{l|cc|cccc|cccc|ccc|c}
        \toprule
        \multirow{2}{*}{Method}
        & \multicolumn{2}{c|}{Ring Stacking}
        & \multicolumn{4}{c|}{Table Sweeping}
        & \multicolumn{4}{c|}{Conveyor Sorting}
        & \multicolumn{3}{c|}{Basket Relocation}
        & \multirow{2}{*}{Avg.} \\

        \cmidrule(lr){2-3}
        \cmidrule(lr){4-7}
        \cmidrule(lr){8-11}
        \cmidrule(lr){12-14}

        & \shortstack{Put\_1}
        & \shortstack{+Put\_2}
        & \shortstack{Sweep\_1}
        & \shortstack{+Sweep\_2}
        & \shortstack{+Lift}
        & \shortstack{+Empty}
        & \shortstack{Pick\_1}
        & \shortstack{+Place\_1}
        & \shortstack{+Pick\_2}
        & \shortstack{+Place\_2}
        & \shortstack{Put Toys}
        & \shortstack{+Grasp}
        & \shortstack{+Place}
        & \\
        \midrule

        OpenVLA-OFT$^\dagger$~\cite{oft}
        & 64\% & 48\%
        & 48\% & 40\% & 24\% & 8\%
        & 32\% & 16\% & 8\% & 4\%
        & 52\% & 24\% & 4\%
        & 16.0\% \\

        $\pi_{0.5}^\dagger$~\cite{pi05}
        & 88\% & 80\%
        & 80\% & 76\% & 64\% & 52\%
        & 64\% & 52\% & 44\% & 44\%
        & 92\% & 72\% & 48\%
        & 56.0\% \\
        
        \rowcolor{tabBg}
        \textbf{VLDA}
        
        & \textbf{96\%} & \textbf{92\%}
        & \textbf{88\%} & \textbf{88\%} & \textbf{80\%} & \textbf{68\%}
        & \textbf{72\%} & \textbf{64\%} & \textbf{60\%} & \textbf{60\%}
        & \textbf{100\%} & \textbf{88\%} & \textbf{72\%}
        & \textbf{73.0\%} \\

        \bottomrule
    \end{tabular}
    }
\vspace{-5pt}
\end{table*}



\section{Experiments}
\subsection{Experimental Setup}
\textbf{Simulation Benchmarks}. We evaluate VLDA on three simulation benchmarks: LIBERO~\cite{libero}, LIBERO-plus~\cite{fei2025libero}, and RoboTwin 2.0~\cite{chen2025robotwin}, covering task generalization, robustness, and cross-distribution generalization. For RoboTwin 2.0, we train on the easy (clean) setting and evaluate on the hard (random) setting. For LIBERO-plus, we directly evaluate the model trained on standard LIBERO in a zero-shot manner. Detailed benchmark descriptions are provided in \textbf{Appendix~A.1}.

\paragraph{Real-World Settings.} We evaluate VLDA and two strong baselines on four long-horizon tasks using the GALAXEA R1 Lite~\cite{galaxea} and AgileX Cobot Magic~\cite{aloha} platforms. We collect 100, 125, 125, and 150 human-teleoperated demonstrations for the four tasks, respectively, and conduct 25 evaluation trials per task and platform. Detailed settings are provided in \textbf{Appendix~A.2}.

\paragraph{Implementation Details.} All experiments are conducted on 8$\times$ NVIDIA A800-SXM4 (80 GB) GPUs, while real-world inference is performed on a single 32 GB consumer GPU. Further model and training details are provided in \textbf{Appendix~A.3}.

\subsection{Simulation Results and Analysis}
\textbf{Performance on LIBERO.} 
As shown in Tab.~\ref{tab:libero_main}, VLDA achieves the best overall performance on LIBERO~\cite{libero}, with an average success rate of \textbf{98.5\%}. By explicitly capturing task-relevant transition cues for action generation while suppressing static and action-irrelevant visual information, its compact actionable Delta consistently improves over the $\pi_{0.5}$ backbone across Spatial (+1.0), Object (+1.4), Goal (+1.4), and Long (\textbf{+2.8}), while surpassing other recent methods to establish a new state of the art.

\paragraph{Performance on LIBERO-plus.} 
As shown in Tab.~\ref{tab:libero_plus_main}, VLDA achieves the best overall zero-shot generalization on LIBERO-Plus \cite{fei2025libero}, with an average success rate of \textbf{84.5\%}, outperforming the $\pi_{0.5}$ backbone (+10.9) and the strongest baseline, GuidedVLA (+9.1). On the more challenging Long suite, the gains increase to \textbf{13.4} and \textbf{12.1} points. These results demonstrate that actionable Delta modeling enhances execution consistency and robustness under distribution shifts by focusing on task-relevant transitions.

\paragraph{Performance on RoboTwin 2.0.} 
As shown in Tab.~\ref{tab:robotwin_main}, VLDA achieves a \textbf{62.9\%} average success rate under zero-shot hard evaluation, outperforming the $\pi_{0.5}$ backbone by \textbf{20.2} points and ranking first across all 10 tasks. The gains are particularly pronounced on complex state-transition tasks, including Put Bottles Dustbin ({+46}), Stacking Three ({+30}), Place Dual Shoes ({+27}), and Open Microwave ({+27}), demonstrating improved robustness to layout, illumination, and appearance shifts through task-relevant Delta modeling.

\begin{table}[t]
\centering
\begin{minipage}[t]{0.49\textwidth}
\centering

\captionof{table}{
Results on the LIBERO benchmark~\cite{libero}.
Ablations: $\alpha$ = \textit{Delta-Seeded Action Flow},
$\beta$ = \textit{Transition Pre-Training},
$\gamma$ = \textit{Full-Future Prediction}.
}
\label{tab:libero_main}

\vspace{-5pt}
\footnotesize
\setlength{\tabcolsep}{3.5pt}
\renewcommand{\arraystretch}{1.05}

\resizebox{\linewidth}{!}{%
\begin{tabular}{l|ccccc}
\toprule
\textbf{Method}
& \textbf{Spatial}
& \textbf{Object}
& \textbf{Goal}
& \textbf{Long}
& \textbf{Avg.} \\
\midrule

InternVLA-M1~\cite{chen2025internvlam1}
& 98.0 & 99.0 & 93.8 & 92.6 & 95.9 \\

gr00t-N1~\cite{bjorck2025gr00t}
& 94.4 & 97.6 & 93.0 & 90.6 & 93.9 \\

OpenVLA~\cite{openvla}
& 84.7 & 88.4 & 79.2 & 53.7 & 76.5 \\

OpenVLA-OFT~\cite{oft}
& 97.6 & 98.4 & 97.9 & 94.5 & 97.1 \\

Diffusion Policy~\cite{diff}
& 78.5 & 87.5 & 73.5 & 64.8 & 76.1 \\

VLA-Adapter~\cite{wang2025vlaadapter}
& 97.8 & 99.2 & 97.2 & 95.0 & 97.3 \\

UniVLA~\cite{univla}
& 95.4 & 98.8 & 93.6 & 94.0 & 95.4 \\

StarVLA-OFT~\cite{starvla}
& 97.8 & 98.6 & 96.2 & 93.8 & 96.6 \\

CronusVLA~\cite{li2025cronusvla}
& 97.3 & 99.6 & 96.9 & 94.0 & 97.0 \\

$\pi_0$~\cite{pi0}
& 96.8 & 98.8 & 95.8 & 85.2 & 94.2 \\

$\pi_{0.5}$~\cite{pi05}
& 98.8 & 98.2 & 98.0 & 92.4 & 96.9 \\

\midrule
\rowcolor{tabBg}

\textbf{VLDA}
& \textbf{99.8}
& \textbf{99.6}
& \textbf{99.4}
& \textbf{95.2}
& \textbf{98.5} \\



~~~~\textit{w/o $\alpha$}
& 98.2 & 98.4 & 98.6 & 93.6 & 97.2 \\

~~~~\textit{w/o $\beta$}
& 98.4 & 98.8 & 98.4 & 93.6 & 97.3 \\

~~~~\textit{w/ $\gamma$}
& 99.0 & 99.0 & 98.6 & 94.6 & 97.8 \\

\bottomrule
\end{tabular}%
}

\end{minipage}
\hfill
\begin{minipage}[t]{0.49\textwidth}
\centering
\captionsetup{skip=3pt}
\captionof{table}{
Ablation of LT-Filter and GT-Anchor on LIBERO and real-world tasks.
}
\label{tab:ablation_2}

\vspace{2pt}

\setlength{\tabcolsep}{2.5pt}
\renewcommand{\arraystretch}{1.35}

\resizebox{\linewidth}{!}{%
\begin{tabular}{cc|cccc|cccc}
\toprule
\multicolumn{2}{c|}{\textbf{Method}}
& \multicolumn{4}{c|}{\textbf{LIBERO}}
& \multicolumn{4}{c}{\textbf{Real-World}} \\
\cmidrule(lr){1-2}
\cmidrule(lr){3-6}
\cmidrule(lr){7-10}

\textbf{LTF}
& \textbf{GTA}
& \textbf{Spatial}
& \textbf{Object}
& \textbf{Goal}
& \textbf{Long}
& \textbf{RS.}
& \textbf{TS.}
& \textbf{CS.}
& \textbf{BR.} \\
\midrule

&
& 98.6 & 98.4 & 98.0 & 92.8
& 76 & 52 & 48 & 48 \\

$\checkmark$
&
& 99.4 & 99.4 & 99.0 & 94.4
& 84 & 60 & 56 & 68 \\

&
$\checkmark$
& 98.8 & 99.0 & 98.8 & 93.8
& 80 & 60 & 52 & 60 \\

\rowcolor{ablBg}
$\checkmark$
& $\checkmark$
& \textbf{99.8}
& \textbf{99.6}
& \textbf{99.4}
& \textbf{95.2}
& \textbf{92}
& \textbf{68}
& \textbf{60}
& \textbf{72} \\

\bottomrule
\end{tabular}%
}

\vspace{2pt}

\captionof{table}{
Ablation study of Delta-to-Action Manifold Grounding on
LIBERO and LIBERO-plus.
}
\label{tab:ablation_3}

\vspace{2pt}

\setlength{\tabcolsep}{3pt}
\renewcommand{\arraystretch}{1.35}

\resizebox{\linewidth}{!}{%
\begin{tabular}{l|c|cccc}
\toprule
\multirow{2}{*}{\textbf{Method}}
& \textbf{LIBERO}
& \multicolumn{4}{c}{\textbf{LIBERO-plus}} \\
\cmidrule(lr){2-2}
\cmidrule(lr){3-6}

& \textbf{Avg.}
& \textbf{Camera}
& \textbf{Light}
& \textbf{Background}
& \textbf{Noise} \\
\midrule

w/o DAMG
& 98.2
& 69.5
& 89.7
& 91.8
& 86.9 \\

\rowcolor{ablBg}
\textbf{VLDA}
& \textbf{98.5}
& \textbf{72.4}
& \textbf{96.0}
& \textbf{95.5}
& \textbf{89.4} \\

\bottomrule
\end{tabular}%
}

\end{minipage}

\vspace{-4pt}
\end{table}

\subsection{Real-world Experimental Analysis}
\label{sec:real_world}

\paragraph{Quantitative Results.}
As shown in Tab.~\ref{tab:real_world_results}, VLDA achieves the best real-world performance at \textbf{73.0\%}, outperforming $\pi_{0.5}$ by \textbf{17.0} points.
Across four tasks, VLDA maintains higher success rates at later stages, demonstrating stronger long-horizon consistency in spatial alignment, tool use, dynamic manipulation, and bimanual coordination. Basket Relocation further tests spatial generalization using randomized mat positions during data collection and OOD target locations at evaluation. While both baselines degrade sharply at the final placement stage, VLDA retains \textbf{72\%} success (\textbf{+24}), showing robust adaptation to unseen layouts.

\begin{figure}[ht]
\centering
\includegraphics[width=\columnwidth]{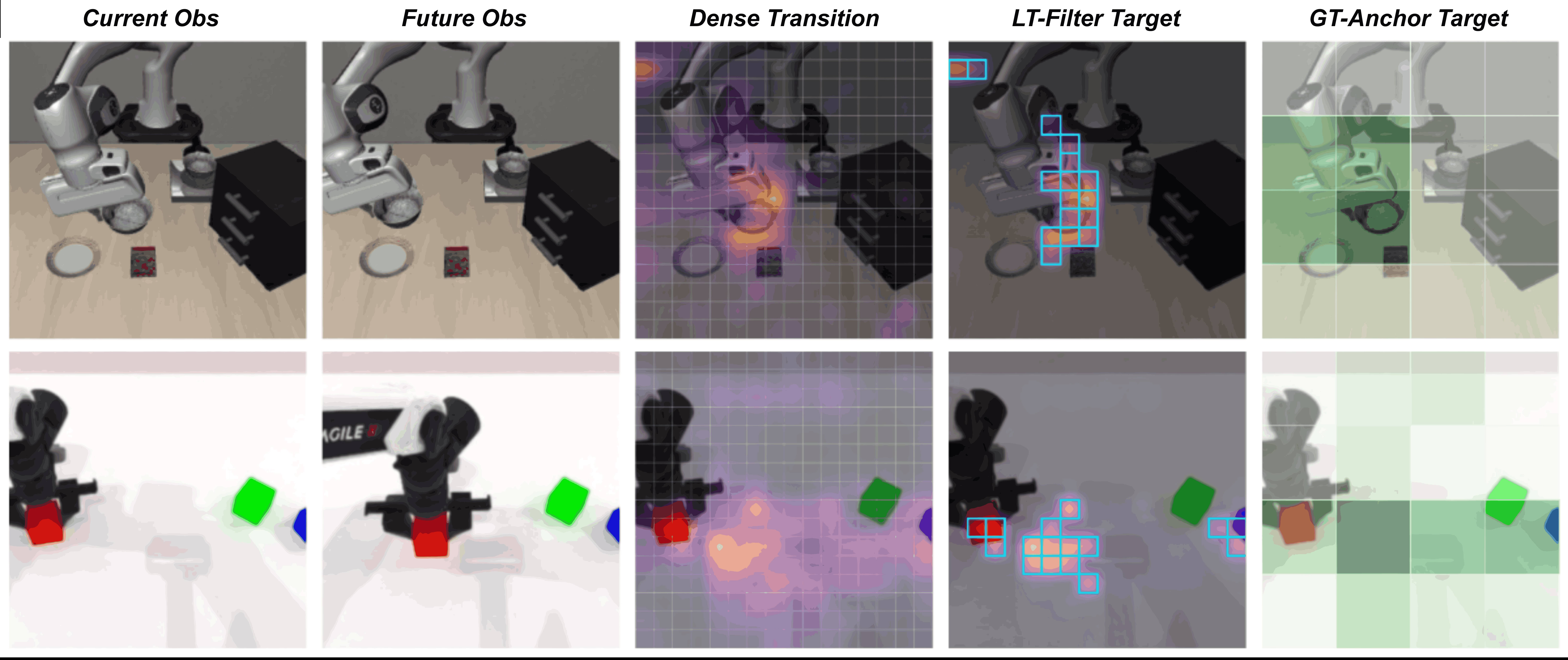}
\vspace{-15pt}
\caption{\textbf{Visualization of Delta supervision.}
LT-Filter extracts sparse manipulation-centric local changes, while GT-Anchor preserves the coarse global transition trend.}
\label{fig:vis2}
\vspace{-10pt}
\end{figure}

\paragraph{Qualitative Visualization.}
Fig.~\ref{fig:vis1} shows executions across four real-world tasks. The current-to-future transitions correspond to manipulation actions, including precise placement, tool-mediated transfer, dynamic interception, and bimanual transport. In Basket Relocation, VLDA adapts to the configuration and two OOD mat locations with horizontal and vertical displacements. These results demonstrate that actionable Delta captures manipulation-relevant local changes while preserving the global direction of the transition.

\vspace{-2pt}

\subsection{Ablation Study}
\paragraph{Ablation of the Transition Paradigm.}
As shown in Tab.~\ref{tab:libero_main}, we examine transition representation and training strategy. \textbf{1)} Replacing Delta modeling with \textit{Full-Future Prediction} reduces performance from 98.5\% to 97.8\%, showing that state-change modeling is a more effective action interface than full-future prediction. \textbf{2)} Jointly optimizing modules from the outset reduces performance to 97.3\%. These results validate compact Delta modeling and staged optimization for establishing a stable, action-oriented transition interface. Details are provided in \textbf{Appendix B.2}.

\vspace{-2pt}

\paragraph{Ablation on LT-Filter and GT-Anchor.}
As shown in Tab.~\ref{tab:ablation_2}, both modules improve performance in simulation and real-world settings, with their combination achieving the best results throughout. 
LT-Filter improves performance by suppressing static and action-irrelevant content, while GT-Anchor provides complementary global transition guidance.
Their clear gain on Basket Relocation further confirms that local transition selectivity and global relocation consistency are both critical for adapting to varying layouts.

\vspace{-2pt}

\paragraph{Ablation on Delta-Seeded Action Flow.} As shown in Tab.~\ref{tab:libero_main}, removing DSAF, including both the Delta-induced noise bias and Delta prefix token, decreases the average success rate from 98.5\% to 97.2\%. This indicates that using the predicted Delta to seed and condition the action flow enables the Action Expert to better exploit the learned transition prior. More detailed ablations are provided in \textbf{Appendix~B.2}.

\vspace{-2pt}

\paragraph{Ablation on Delta-to-Action Manifold Grounding.}
As shown in Tab.~\ref{tab:ablation_3}, Manifold Grounding consistently improves performance on standard LIBERO, with its benefits becoming more pronounced under LIBERO-plus perturbations that alter camera, lighting, background, and noise while preserving task semantics. 
The gains are especially substantial under lighting (\textbf{+6.3}) and background (\textbf{+3.7}) shifts, indicating that Manifold Grounding steers Delta toward action-relevant transitions while suppressing action-irrelevant visual variations.

\vspace{-2pt}

\subsection{Visualizing Delta Supervision}
\label{sec:delta_visualization}
As shown in Fig.~\ref{fig:vis2}, the first three columns present the current observation, future observation, and their dense feature Delta (future minus current), where higher-intensity colors indicate changes. 
The subsequent columns show that LT-Filter selects sparse manipulation-centric local changes around the target object and interaction regions, while GT-Anchor preserves a coarse scene-level summary of the future transition. Together, they provide complementary local selectivity and global consistency for Delta learning. More visualizations and analyses are provided in \textbf{Appendix~C.2}.

\vspace{-2pt}
\section{Conclusion}
We propose Vision-Language-Delta-Action (VLDA), which models compact actionable transitions instead of direct actions or complete future states. LT-Filter captures manipulation-relevant changes; GT-Anchor preserves global transition consistency; DDAR grounds Delta and guides action refinement. Experiments demonstrate improved long-horizon consistency, robustness, and spatial generalization, validating actionable Delta as an effective interface between vision-language understanding and action execution.

\bibliographystyle{plainnat}
\bibliography{references}

\clearpage












\title{Supplementary Material}




\maketitle

\appendix

We provide comprehensive supplementary material to support the methodology, implementation, and analysis introduced in the main paper on VLDA. The content is organized as follows:
\begin{itemize}
    \setlength{\itemsep}{0.3em}
    \setlength{\parsep}{0pt}
    
    \item \textbf{Section~\ref{app:experimental_details}: Experimental Details}
    \\[0.3em] \hspace*{2em}A.1 Simulation Benchmarks
    \\[0.3em] \hspace*{2em}A.2 Real-world Platforms and Tasks
    \\[0.3em] \hspace*{2em}A.3 Model and Training Details
    
    \item \textbf{Section~\ref{app:section_b}: Extended Quantitative Analyses}
    \\[0.3em] \hspace*{2em}B.1 Extended Platform Results
    \\[0.3em] \hspace*{2em}B.2 Extended Ablation Results

    \item \textbf{Section~\ref{app:section_c}: Extended Qualitative Analyses}
    \\[0.3em] \hspace*{2em}C.1 Real-World Task Execution Visualization
    \\[0.3em] \hspace*{2em}C.2 Delta Supervision Visualization

    \item \textbf{Section~\ref{app:section_d}: Discussion}
    \\[0.3em] \hspace*{2em}D.1 Collapse Mitigation in Transition Representation Learning
    \\[0.3em] \hspace*{2em}D.2 Entropy-Regularized Set Distance
    \\[0.3em] \hspace*{2em}D.3 Supplementary Details on Motivation
    \\[0.3em] \hspace*{2em}D.4 Mathematical Details of Key Equations
    \\[0.3em] \hspace*{2em}D.5 Future Work and Limitations
\end{itemize}
We also provide a multimedia video as an additional supplementary material. This video offers dynamic demonstrations of the proposed methodology and execution pipeline, including intuitive illustrations of the key components and detailed real-world robotic manipulation results.

\section{Experimental Details}
\label{app:experimental_details}

\subsection{Simulation Benchmarks}
\label{app:simulation_benchmarks}

\paragraph{LIBERO.}
LIBERO~\cite{libero} is a language-conditioned robotic manipulation benchmark
designed to evaluate the transfer of declarative and procedural
knowledge across tasks. We adopt the four commonly used evaluation
suites: Spatial, Object, Goal, and Long, each containing 10 tasks.
The Spatial suite requires distinguishing objects according to their
relative positions under otherwise similar scene configurations.
The Object suite varies the target object while retaining comparable
manipulation behaviors and scene layouts. The Goal suite uses similar
objects and spatial configurations but changes the desired task goal,
thereby emphasizing the transfer of motion and behavioral knowledge.
The Long suite consists of compositional, multi-stage tasks that
require extended temporal reasoning and coordinated execution.

\paragraph{RoboTwin 2.0.}
RoboTwin 2.0~\cite{chen2025robotwin} is a simulation benchmark designed for generalizable
bimanual manipulation. It provides 50 dual-arm tasks involving
diverse objects, scene configurations, and coordinated manipulation
skills. In addition to clean environments, the benchmark supports
domain-randomized evaluation with variations in scene clutter,
illumination, background textures, and tabletop height. These
variations jointly alter both visual appearance and object geometry,
making the benchmark particularly suitable for evaluating whether a
policy can preserve bimanual coordination and task execution under
substantial distribution shifts. Under our clean-to-randomized
protocol, performance therefore reflects the ability to transfer
manipulation knowledge beyond the narrow visual and spatial
distribution observed during training.

\paragraph{LIBERO-Plus.}
LIBERO-Plus~\cite{fei2025libero} extends the original LIBERO environments with controlled
perturbations along seven dimensions: object layouts, camera
viewpoints, robot initial states, language instructions, lighting
conditions, background textures, and sensor noise. It contains
10,030 perturbed task instances organized into five difficulty
levels. In contrast to the broad scene randomization used in
RoboTwin 2.0, LIBERO-Plus enables a factor-wise diagnosis of model
robustness by isolating individual sources of distribution shift,
while also supporting evaluation under combined perturbations.
Since no LIBERO-Plus-specific adaptation is performed in our
experiments, this benchmark measures whether the representations
learned from standard LIBERO remain reliable under unseen changes
in viewpoint, robot configuration, visual appearance, language, and
observation quality.

\begin{table*}[t]
    \centering
    \caption{Real-world Task Details. 
    Language instructions and progressive evaluation checkpoints for
    the four real-world tasks. Checkpoints are cumulative: reaching
    checkpoint $C_k$ requires successful completion of all preceding
    checkpoints.
    }
    \label{tab:real_task_protocol}
    \small
    \setlength{\tabcolsep}{4.2pt}
    \renewcommand{\arraystretch}{1.18}
    \begin{tabular}{p{0.05\textwidth} p{0.10\textwidth} p{0.10\textwidth} p{0.26\textwidth} p{0.40\textwidth}}
        \toprule
        Task & Name & Number & Language Instruction & Progressive Checkpoints \\
        \midrule

        1 &
        Ring Stacking &
        200 (100$\times2$) &
        \emph{Stack the yellow and purple rings on the peg.} &
        $C_1$: the yellow ring is placed around the peg;
        $C_2$: the purple ring is placed above the yellow ring. \\

        2 &
        Conveyor Sorting &
        250 (125$\times2$) &
        \emph{Pick the cup and bun from the conveyor and place them
        on the plate.} &
        $C_1$: the moving cup is successfully grasped;
        $C_2$: the cup is placed upright perfectly on the plate;
        $C_3$: the moving bun is successfully grasped;
        $C_4$: the bun is placed on the plate while distractor objects
        remain unselected. \\

        3 &
        Table Sweeping &
        250 (125$\times2$) &
        \emph{Sweep the paper balls into the dustpan and empty them
        onto the plate.} &
        $C_1$: the first paper ball is swept into the dustpan;
        $C_2$: both paper balls are inside the dustpan;
        $C_3$: the dustpan is lifted without losing either paper ball;
        $C_4$: both paper balls are emptied onto the plate. \\

        4 &
        Basket Relocation &
        300 (150$\times2$) &
        \emph{Put both toys in the basket and move the basket onto the
        blue mat.} &
        $C_1$: both toys are placed inside the basket;
        $C_2$: the basket is securely grasped with both arms;
        $C_3$: the basket is placed on the blue mat without
        dropping either toy. \\

        \bottomrule
    \end{tabular}
\end{table*}

\begin{figure}[ht]
\centering
\includegraphics[width=0.6\columnwidth]{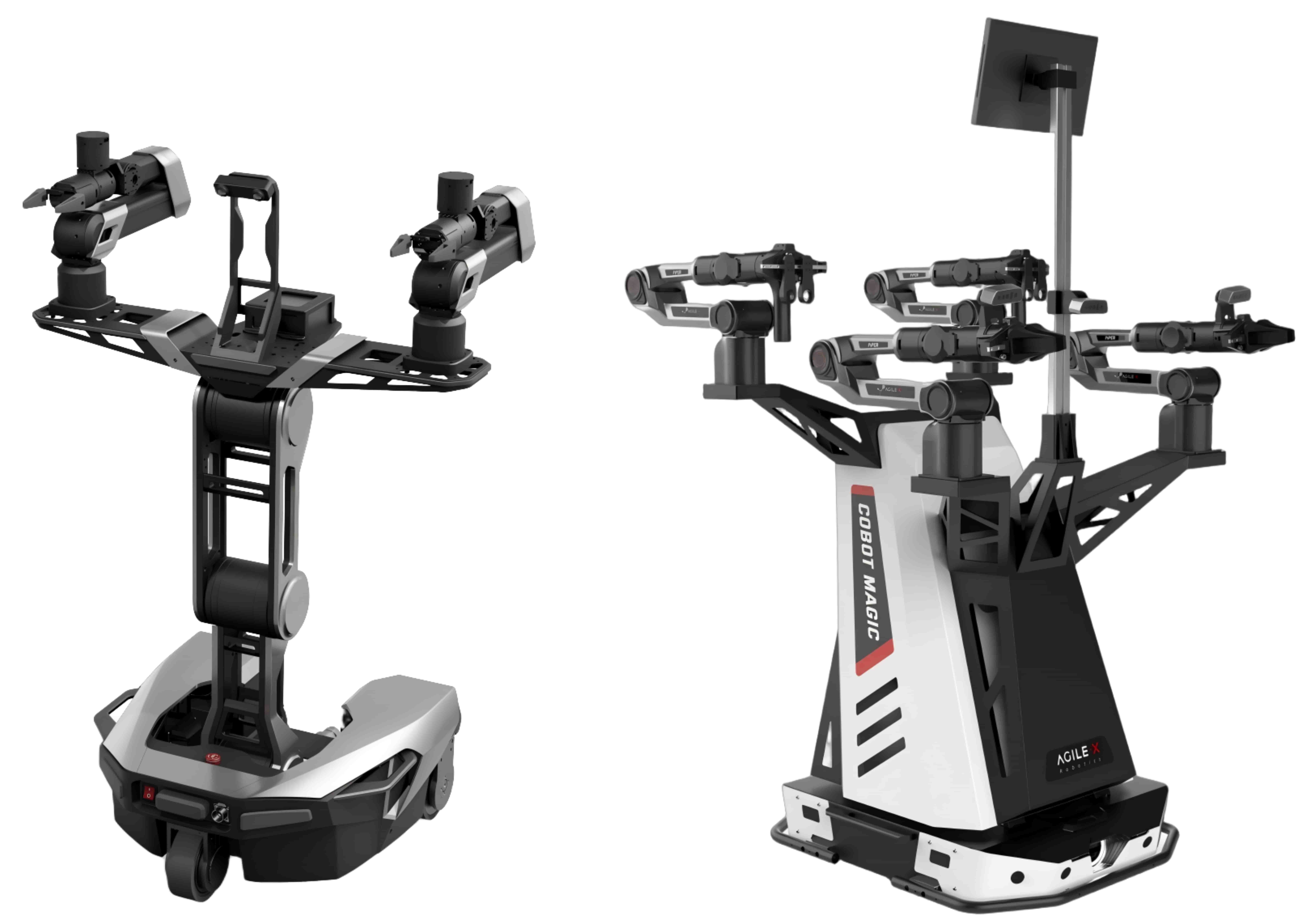}
\vspace{-3pt}
\caption{GALAXEA R1 Lite (left) and AgileX Cobot Magic (right) platforms.}
\label{fig:supp-robot}
\end{figure}

\subsection{Real-world Platforms and Tasks}

\paragraph{GALAXEA R1 Lite Platform.}
As shown on the left of Fig.~\ref{fig:supp-robot}, the GALAXEA R1 Lite~\cite{galaxea} is a 23-DoF mobile bimanual manipulation platform, comprising a wheeled chassis, an articulated torso, and two 6-DoF robotic arms equipped with parallel grippers. The movable base and torso extend the effective workspace of the robot, while platform-level and wrist-mounted cameras provide complementary global and close-range visual observations. We employ this platform to evaluate VLDA under large-workspace and coordinated bimanual manipulation settings.

\paragraph{AgileX Cobot Magic Platform.}
As shown on the right of Fig.~\ref{fig:supp-robot}, the AgileX Cobot Magic platform~\cite{aloha} is a low-cost, open-source bimanual system developed upon Stanford's Mobile ALOHA project for teleoperated data collection and imitation learning. It integrates a Tracer differential-drive mobile base, two 6-DoF follower arms with parallel-jaw grippers, and RGB-D sensors. During demonstration collection, the follower arms are controlled through a pair of leader arms, while fixed and wrist-mounted RGB cameras provide multi-view observations of the workspace and end-effector interactions. Compared with R1 Lite, ALOHA is operated in a stationary tabletop configuration, emphasizing precise, contact-rich, and tightly coordinated bimanual manipulation.

\paragraph{Real-world Task Details.}
We design four long-horizon real-world tasks covering sequential spatial
alignment, contact-rich tool use, dynamic target interception, and
bimanual object relocation. We collected human-teleoperated demonstrations from individual platforms, with corresponding sample counts of 100, 125, 125, and 150 in order. OpenVLA-OFT, $\pi_{0.5}$, and VLDA are trained using the same demonstrations, observations, robot states, and control frequency to ensure a fair comparison.

The language instructions specify only the task objective and necessary
semantic order, without exposing low-level actions or evaluation
checkpoints. For Basket Relocation, the blue mat is randomly positioned
during data collection, and multiple target locations are used during
evaluation to assess spatial generalization. Each method is evaluated
over 25 physical trials per task and platform. The instructions and
progressive checkpoints are summarized in
Tab.~\ref{tab:real_task_protocol}.

\paragraph{Evaluation Protocol.}
For each method, task, and platform, we conduct 25 independent physical trials. At the beginning of each trial, task-relevant objects are randomly perturbed within predefined regions while preserving the task semantics. For Conveyor Sorting, the conveyor speed and object release interval are fixed across methods, while the ordering and spacing of target and distractor objects are varied. For Basket Relocation, the blue mat is placed at multiple target locations to evaluate spatial generalization.

Success is recorded at each progressive checkpoint. A trial reaches checkpoint $C_k$ only if all preceding checkpoints are successfully completed, and complete-task success is defined by the final checkpoint. All results are reported as success rates over 25 trials, while the average success rate is computed from the complete-task success rates across the four tasks.

\begin{table*}[htbp]
    \centering
    \caption{
    Per-task success rates on the 50 RoboTwin 2.0 tasks under
    the Randomized setting. 
    }
    \label{tab:robotwin_random_results}
    \setlength{\tabcolsep}{4.5pt}
    \renewcommand{\arraystretch}{1.02}
    \resizebox{\textwidth}{!}{%
    \begin{tabular}{l|cccccccccc}
        \toprule
        \multirow{1}{*}{\textit{Task}}
        & ACT
        & DP
        & DP3
        & RDT-1B
        & $\pi_0$
        & $\pi_{0.5}$
        & TwinVLA
        & BagelVLA
        & HALO
        & \textbf{VLDA} \\
        \midrule
        \textit{Adjust Bottle}
        & 23\% & 0\% & 3\% & 75\% & 56\% & \textbf{83\%} & 35\% & 14\% & 9\% & 76\% \\

        \textit{Beat Block Hammer}
        & 3\% & 0\% & 8\% & 37\% & 21\% & 50\% & 10\% & 16\% & 11\% & \textbf{52\%} \\

        \textit{Blocks Ranking RGB}
        & 0\% & 0\% & 0\% & 0\% & 5\% & 35\% & 0\% & 4\% & 7\% & \textbf{53\%} \\

        \textit{Blocks Ranking Size}
        & 0\% & 0\% & 0\% & 0\% & 1\% & 14\% & 0\% & 2\% & 2\% & \textbf{16\%} \\

        \textit{Click Alarmclock}
        & 4\% & 5\% & 14\% & 12\% & 11\% & \textbf{93\%} & 1\% & 20\% & 14\% & 64\% \\

        \textit{Click Bell}
        & 3\% & 0\% & 0\% & 9\% & 3\% & \textbf{76\%} & 13\% & 35\% & 10\% & 67\% \\

        \textit{Dump Bin Bigbin}
        & 1\% & 0\% & 53\% & 32\% & 24\% & 42\% & 34\% & 51\% & 28\% & \textbf{73\%} \\

        \textit{Grab Roller}
        & 25\% & 0\% & 2\% & 43\% & 80\% & 89\% & 22\% & 41\% & 57\% & \textbf{96\%} \\

        \textit{Handover Block}
        & 0\% & 0\% & 0\% & 14\% & 8\% & 19\% & 0\% & 0\% & \textbf{36\%} & 22\% \\

        \textit{Handover Mic}
        & 0\% & 0\% & 3\% & 31\% & 13\% & 18\% & 2\% & 8\% & \textbf{61\%} & 38\% \\

        \textit{Hanging Mug}
        & 0\% & 0\% & 1\% & \textbf{16\%} & 3\% & 3\% & 5\% & 1\% & 5\% & 12\% \\

        \textit{Lift Pot}
        & 0\% & 0\% & 0\% & 9\% & 36\% & 0\% & 7\% & 32\% & 34\% & \textbf{72\%} \\

        \textit{Move Can Pot}
        & 4\% & 0\% & 6\% & 12\% & 21\% & \textbf{27\%} & 5\% & 0\% & 15\% & 26\% \\

        \textit{Move Pillbottle Pad}
        & 0\% & 0\% & 0\% & 0\% & 1\% & 29\% & 2\% & 1\% & 26\% & \textbf{40\%} \\

        \textit{Move Playingcard Away}
        & 0\% & 0\% & 3\% & 11\% & 22\% & 67\% & 35\% & 30\% & 53\% & \textbf{68\%} \\

        \textit{Move Stapler Pad}
        & 0\% & 0\% & 0\% & 0\% & 2\% & 18\% & 0\% & 0\% & \textbf{19\%} & 16\% \\

        \textit{Open Laptop}
        & 0\% & 0\% & 7\% & 32\% & 46\% & 35\% & 17\% & 37\% & 37\% & \textbf{61\%} \\

        \textit{Open Microwave}
        & 0\% & 0\% & 22\% & 20\% & 50\% & 37\% & 1\% & 0\% & 24\% & \textbf{64\%} \\

        \textit{Pick Diverse Bottles}
        & 0\% & 0\% & 1\% & 0\% & 6\% & 3\% & 8\% & \textbf{34\%} & 17\% & \textbf{34\%} \\

        \textit{Pick Dual Bottles}
        & 0\% & 0\% & 1\% & 13\% & 12\% & 6\% & 12\% & \textbf{56\%} & 30\% & 39\% \\

        \textit{Place A2B Left}
        & 0\% & 0\% & 2\% & 1\% & 1\% & \textbf{60\%} & 5\% & 12\% & 8\% & 45\% \\

        \textit{Place A2B Right}
        & 0\% & 0\% & 0\% & 1\% & 6\% & \textbf{57\%} & 1\% & 11\% & 9\% & 46\% \\

        \textit{Place Bread Basket}
        & 0\% & 0\% & 1\% & 2\% & 4\% & \textbf{56\%} & 3\% & 29\% & 26\% & 54\% \\

        \textit{Place Bread Skillet}
        & 0\% & 0\% & 0\% & 1\% & 1\% & \textbf{46\%} & 1\% & 26\% & 23\% & 42\% \\

        \textit{Place Burger Fries}
        & 0\% & 0\% & 18\% & 27\% & 4\% & \textbf{70\%} & 13\% & 11\% & 37\% & 64\% \\

        \textit{Place Can Basket}
        & 0\% & 0\% & 2\% & 6\% & 5\% & 25\% & 0\% & 0\% & \textbf{34\%} & 16\% \\

        \textit{Place Cans Plasticbox}
        & 0\% & 0\% & 3\% & 5\% & 2\% & \textbf{47\%} & 8\% & 5\% & \textbf{47\%} & 36\% \\

        \textit{Place Container Plate}
        & 1\% & 0\% & 1\% & 17\% & 45\% & 78\% & 4\% & 58\% & 22\% & \textbf{79\%} \\

        \textit{Place Dual Shoes}
        & 0\% & 0\% & 0\% & 4\% & 0\% & 7\% & 3\% & 0\% & 3\% & \textbf{34\%} \\

        \textit{Place Empty Cup}
        & 0\% & 0\% & 1\% & 7\% & 11\% & \textbf{86\%} & 1\% & 34\% & 28\% & 79\% \\

        \textit{Place Fan}
        & 0\% & 0\% & 1\% & 2\% & 10\% & 36\% & 0\% & 5\% & 9\% & \textbf{38\%} \\

        \textit{Place Mouse Pad}
        & 0\% & 0\% & 1\% & 0\% & 1\% & 26\% & 0\% & 14\% & 12\% & \textbf{29\%} \\

        \textit{Place Object Basket}
        & 0\% & 0\% & 0\% & 17\% & 2\% & 36\% & 3\% & 3\% & 25\% & \textbf{41\%} \\

        \textit{Place Object Scale}
        & 0\% & 0\% & 0\% & 0\% & 0\% & \textbf{49\%} & 0\% & 0\% & 5\% & 45\% \\

        \textit{Place Object Stand}
        & 0\% & 0\% & 0\% & 5\% & 11\% & 65\% & 2\% & 21\% & 33\% & \textbf{67\%} \\

        \textit{Place Phone Stand}
        & 0\% & 0\% & 2\% & 6\% & 7\% & 53\% & 2\% & 2\% & 10\% & \textbf{57\%} \\

        \textit{Place Shoe}
        & 0\% & 0\% & 2\% & 7\% & 6\% & 61\% & 4\% & 29\% & 18\% & \textbf{68\%} \\

        \textit{Press Stapler}
        & 6\% & 0\% & 3\% & 24\% & 29\% & \textbf{70\%} & 26\% & 58\% & 64\% & 63\% \\

        \textit{Put Bottles Dustbin}
        & 1\% & 0\% & 21\% & 4\% & 13\% & 9\% & 4\% & 10\% & 13\% & \textbf{55\%} \\

        \textit{Put Object Cabinet}
        & 0\% & 0\% & 1\% & \textbf{18\%} & \textbf{18\%} & 15\% & 16\% & 0\% & 8\% & 11\% \\

        \textit{Rotate QRCode}
        & 0\% & 0\% & 1\% & 5\% & 15\% & \textbf{56\%} & 3\% & 21\% & 11\% & 49\% \\

        \textit{Scan Object}
        & 0\% & 0\% & 1\% & 1\% & 1\% & \textbf{38\%} & 4\% & 32\% & 24\% & 20\% \\

        \textit{Shake Bottle Horizontally}
        & 4\% & 18\% & 25\% & 51\% & 51\% & \textbf{100\%} & 55\% & 73\% & 66\% & \textbf{100\%} \\

        \textit{Shake Bottle}
        & 10\% & 8\% & 19\% & 45\% & 60\% & \textbf{100\%} & 58\% & 74\% & 73\% & 99\% \\

        \textit{Stack Blocks Three}
        & 0\% & 0\% & 0\% & 0\% & 0\% & 16\% & 0\% & 5\% & \textbf{37\%} & 32\% \\

        \textit{Stack Blocks Two}
        & 0\% & 0\% & 0\% & 2\% & 1\% & 56\% & 0\% & 6\% & \textbf{60\%} & 53\% \\

        \textit{Stack Bowls Three}
        & 0\% & 0\% & 5\% & 17\% & 24\% & 35\% & 15\% & 13\% & 25\% & \textbf{65\%} \\

        \textit{Stack Bowls Two}
        & 0\% & 0\% & 6\% & 30\% & 41\% & 66\% & 11\% & 52\% & 49\% & \textbf{86\%} \\

        \textit{Stamp Seal}
        & 0\% & 0\% & 0\% & 0\% & 4\% & 23\% & 1\% & 8\% & 21\% & \textbf{43\%} \\

        \textit{Turn Switch}
        & 2\% & 1\% & 8\% & 15\% & 23\% & 6\% & 15\% & 30\% & 27\% & \textbf{38\%} \\
        \midrule
        \textbf{\textit{Average}}
        & 1.74\% & 0.64\% & 4.96\% & 13.72\% & 16.34\%
        & 43.84\% & 9.34\% & 20.48\% & 26.44\% & \textbf{50.86\%} \\
        \bottomrule
    \end{tabular}%
    }
\end{table*}

\begin{table*}[th]
    \centering
    \caption{
    Real-world task results on the \textbf{ALOHA}~\cite{aloha}. ``$\dagger$'' denotes results reproduced using the same settings as VLDA. Each entry reports the progressive-stage success rate over 25 trials. \textbf{Avg.} is the average complete-task success rate, calculated from the final checkpoint of each task.
    }
    \label{tab:real_world_aloha}
    \scriptsize
    \setlength{\tabcolsep}{1.5pt}
    \resizebox{\textwidth}{!}{
    \begin{tabular}{l|cc|cccc|cccc|ccc|c}
        \toprule
        \multirow{2}{*}{Method}
        & \multicolumn{2}{c|}{Ring Stacking}
        & \multicolumn{4}{c|}{Table Sweeping}
        & \multicolumn{4}{c|}{Conveyor Sorting}
        & \multicolumn{3}{c|}{Basket Relocation}
        & \multirow{2}{*}{Avg.} \\

        \cmidrule(lr){2-3}
        \cmidrule(lr){4-7}
        \cmidrule(lr){8-11}
        \cmidrule(lr){12-14}

        & \shortstack{Put\_1}
        & \shortstack{+Put\_2}
        & \shortstack{Sweep\_1}
        & \shortstack{+Sweep\_2}
        & \shortstack{+Lift}
        & \shortstack{+Empty}
        & \shortstack{Pick\_1}
        & \shortstack{+Place\_1}
        & \shortstack{+Pick\_2}
        & \shortstack{+Place\_2}
        & \shortstack{Put Toys}
        & \shortstack{+Grasp}
        & \shortstack{+Place}
        & \\
        \midrule

        OpenVLA-OFT$^\dagger$~\cite{oft}
        & 60\% & 44\%
        & 48\% & 44\% & 28\% & 12\%
        & 32\% & 16\% & 8\% & 4\%
        & 56\% & 28\% & 8\%
        & 17.0\% \\

        $\pi_{0.5}^\dagger$~\cite{pi05}
        & 88\% & 80\%
        & 80\% & 72\% & 60\% & 48\%
        & 64\% & 52\% & 44\% & 44\%
        & 92\% & 68\% & 44\%
        & 54.0\% \\

        \textbf{VLDA}
        & \textbf{96\%} & \textbf{88\%}
        & \textbf{92\%} & \textbf{88\%} & \textbf{80\%} & \textbf{68\%}
        & \textbf{76\%} & \textbf{68\%} & \textbf{64\%} & \textbf{64\%}
        & \textbf{100\%} & \textbf{88\%} & \textbf{68\%}
        & \textbf{72.0\%} \\

        \bottomrule
    \end{tabular}
    }
\end{table*}

\subsection{Model and Training Details}
\label{app:training_details}

\paragraph{Architecture and Initialization.}
VLDA is implemented on top of $\pi_{0.5}$, which consists of a
PaliGemma-initialized 2B vision-language backbone and a Gemma-300M
Action Expert. We additionally introduce a World Expert with the same
transformer architecture and model capacity as the Action Expert. The
World Expert is initialized independently from scratch rather than
from the pretrained Action Expert.

For visual transition modeling, we adopt the pretrained V-JEPA 2
ViT-g/16 encoder (\texttt{vjepa2\_vit\_giant})~\cite{vjepa2}. The
online and teacher encoders are initialized from the encoder and EMA
target-encoder branches of the pretrained V-JEPA 2 checkpoint,
respectively. Both encoders process the primary-camera image at a
resolution of $256\times256$, producing $P=256$ spatial tokens with
$d_v=1408$. The primary-view restriction applies only to transition
supervision, while the policy backbone continues to process all
available camera views at $224\times224$ resolution.

The online encoder is optimized by gradients, whereas the teacher
encoder remains gradient-free and is updated exclusively through EMA
with momentum $m=0.99925$. This asymmetric design provides a slowly
evolving future-feature target throughout training.

\paragraph{Delta and Projection Modules.}
The World Expert predicts $K=64$ Delta tokens with
$d_\Delta=256$. The down-projection $g_{\downarrow}$ and
up-projection $g_{\uparrow}$ are implemented as two-layer MLPs with
a hidden dimension of 1024. The action-space bias projection $q_b$
uses the same hidden dimension. Its output layer is zero-initialized,
and the unconstrained learnable scale $\gamma$ is initialized to
$0.01$.

The mean-pooled Delta is linearly projected into a 1024-dimensional
prefix token for the Action Expert. For Delta-action alignment,
$q_\Delta$ and $q_{\mathrm{act}}$ each consist of LayerNorm followed
by a linear projection into a shared 256-dimensional contrastive
space.

\paragraph{Two-Stage Training.}
VLDA is optimized in two stages. During the first stage, termed
transition pre-training, we freeze the vision-language backbone and
Action Expert, while optimizing the online V-JEPA 2 encoder, World
Expert, and transition projection modules. This stage is trained for
10K optimization steps with a global batch size of 48.

The second stage is initialized from the transition-pretraining
checkpoint. We unfreeze the vision-language backbone and Action
Expert, and jointly optimize all gradient-enabled modules for an
additional 60K steps with a global batch size of 128. The World Expert,
online encoder, and projection modules remain trainable, while the
teacher encoder continues to receive no gradient and is updated solely
through EMA.

The LT-Filter and GT-Anchor objectives are equally weighted. We set
$\lambda_{(1)}=1$ and $\lambda_{\mathrm{DDAR}}=0.1$ in the
second-stage objective. The Delta-token pooling temperature and
contrastive temperature are set to $\tau_\Delta=0.1$ and
$\tau_c=0.07$, respectively.

Both training stages are conducted on a single machine equipped with 8 NVIDIA A800-SXM4 GPUs, each with 80 GB of memory.

\paragraph{Temporal Alignment and Action Sampling.}
We use an action horizon of $H=10$ for the LIBERO-based experiments
and $H=50$ for RoboTwin 2.0. The future observation used to construct
transition supervision is aligned with the end of the corresponding
action chunk. Near the end of a trajectory, the future index is clipped
to the final available frame:
\begin{equation}
    t_{\mathrm{future}}
    =
    \min(t+H,T_{\mathrm{end}}).
\end{equation}

For Delta-to-Action Manifold Grounding, the ground-truth actions in
the current action chunk are treated as positive samples, with the
number of positive actions capped at 10. Accordingly, all $H$ actions
are used when $H\leq10$, whereas the first 10 actions are used for
longer action chunks. The number of negative actions is set to four
times the number of positive actions. These negatives are sampled
uniformly without replacement from non-overlapping temporal regions
of the same trajectory outside the current action interval.

\paragraph{Inference and Deployment.}
During inference, the World Expert generates the Delta-token set using
three explicit Euler integration steps. The resulting Delta tokens are
mean-pooled and used to construct both the Delta-induced action-noise
bias and the Delta prefix token. Conditioned on these two signals, the
Action Expert generates the action chunk using ten explicit Euler
integration steps.

The two V-JEPA 2 encoders and the modules used exclusively for
transition-target construction and Delta-action alignment, including
$g_{\downarrow}$, $g_{\uparrow}$, $q_\Delta$, and
$q_{\mathrm{act}}$, are discarded after training. Deployment therefore
requires only the vision-language backbone, World Expert,
Delta-to-action conditioning modules, and Action Expert.

\begin{table*}[t]
\caption{Results on the LIBERO-Plus benchmark~\cite{fei2025libero}.
We report zero-shot generalization performance across seven
perturbation dimensions and four task suites using 10,030 total
rollouts. The last three rows compare VLDA with transition-paradigm
and training-strategy ablations. All three variants are trained only
on standard LIBERO and evaluated on LIBERO-Plus without further
adaptation.}
\vspace{-5pt}
\centering
\footnotesize
\setlength{\tabcolsep}{4pt}
\label{tab:libero_plus_ablation}
\resizebox{\textwidth}{!}{
    \begin{tabular}{l|ccccccc|cccc|c}
        \toprule
        \multirow{2}{*}{\textbf{Method}} 
        & \multicolumn{7}{c|}{\textbf{Perturbation Dimensions}}
        & \multicolumn{4}{c|}{\textbf{Task Suites}} & \multirow{2}{*}{\textbf{Avg.}} \\
        \cmidrule(lr){2-8} \cmidrule(lr){9-12}
        & Camera & Robot & Language & Light & Background & Noise & Layout
        & Spatial & Object & Goal & Long &  \\
        \midrule

        w/o Transition Pre-Training
        & 66.6 & 71.5 & 80.4 & 87.7 & 89.4 & 84.3 & 81.6
        & 83.9 & 87.2 & 74.7 & 72.2 & 79.4 \\
        
        w/ Full-Future Prediction
        & 68.0 & 73.3 & 81.8 & 89.1 & 90.8 & 85.5 & 84.4
        & 85.4 & 88.0 & 76.0 & 75.2 & 81.1 \\

        \textbf{VLDA}
        & \textbf{72.4}
        & \textbf{75.3}
        & \textbf{84.0}
        & \textbf{96.0}
        & \textbf{95.5}
        & \textbf{89.4}
        & \textbf{85.7}
        & \textbf{89.6}
        & \textbf{89.8}
        & \textbf{80.1}
        & \textbf{78.3}
        & \textbf{84.5} \\

        \bottomrule
    \end{tabular}
}
\vspace{-5pt}
\end{table*}

\begin{table}[t]
\centering

\begin{minipage}[t]{0.43\textwidth}
\centering

\captionof{table}{
Ablation of the two Delta-conditioning branches in
Delta-Seeded Action Flow on LIBERO.
}
\label{tab:ablation_dsaf}

\vspace{-10pt}
\footnotesize
\setlength{\tabcolsep}{3pt}
\renewcommand{\arraystretch}{1.05}

\resizebox{\linewidth}{!}{%
\begin{tabular}{cc|cccc|c}
\toprule
\textbf{Bias}
&
\textbf{Prefix}
&
\textbf{Spatial}
&
\textbf{Object}
&
\textbf{Goal}
&
\textbf{Long}
&
\textbf{Avg.} \\
\midrule

             &              & 98.2 & 98.4 & 98.6 & 93.6 & 97.2 \\
$\checkmark$ &              & 99.2 & 99.0 & 98.8 & 94.6 & 97.9 \\
             & $\checkmark$ & 98.8 & 98.6 & 98.8 & 94.0 & 97.6 \\
$\checkmark$ & $\checkmark$
& \textbf{99.8}
& \textbf{99.6}
& \textbf{99.4}
& \textbf{95.2}
& \textbf{98.5} \\

\bottomrule
\end{tabular}%
}

\vspace{-8pt}

\captionof{table}{
Ablation of Delta injection strategies for action-flow
initialization on LIBERO.
}
\label{tab:ablation_bias}

\footnotesize
\setlength{\tabcolsep}{3pt}
\renewcommand{\arraystretch}{1.08}

\resizebox{\linewidth}{!}{%
\begin{tabular}{l|cccc|c}
\toprule
\textbf{Method}
&
\textbf{Spatial}
&
\textbf{Object}
&
\textbf{Goal}
&
\textbf{Long}
&
\textbf{Avg.} \\
\midrule

No shift
& 98.8 & 98.6 & 98.8 & 94.0 & 97.6 \\

Direct injection
& 99.2 & 99.0 & 98.8 & 94.8 & 98.0 \\

Proposed bias
& \textbf{99.8}
& \textbf{99.6}
& \textbf{99.4}
& \textbf{95.2}
& \textbf{98.5} \\

\bottomrule
\end{tabular}%
}

\end{minipage}
\hfill
\begin{minipage}[t]{0.55\textwidth}
\centering

\captionof{table}{
Ablations of transition-token selection, set-matching
objectives, and Delta-token number on LIBERO. Within each group, only
the indicated factor is changed.
}
\label{tab:ablation_for_3}

\vspace{-3pt}
\footnotesize
\setlength{\tabcolsep}{2.5pt}
\renewcommand{\arraystretch}{1.06}

\resizebox{\linewidth}{!}{%
\begin{tabular}{ll|cccc|c}
\toprule
\textbf{Factor}
&
\textbf{Variant}
&
\textbf{Spatial}
&
\textbf{Object}
&
\textbf{Goal}
&
\textbf{Long}
&
\textbf{Avg.} \\
\midrule

\multirow{3}{*}{\shortstack{Token\\selection}}
& Random-$K$
& 98.8 & 98.6 & 98.2 & 94.0 & 97.4 \\

& Spatial pooling
& 99.4 & 99.0 & 98.8 & 94.4 & 97.9 \\

& Top-$K$ (ours)
& \textbf{99.8}
& \textbf{99.6}
& \textbf{99.4}
& \textbf{95.2}
& \textbf{98.5} \\

\midrule

\multirow{3}{*}{\shortstack{Set\\matching}}
& Point-wise MSE
& 98.4 & 98.4 & 98.2 & 93.8 & 97.2 \\

& Chamfer distance
& 99.2 & 99.0 & 98.6 & 94.4 & 97.8 \\

& OT matching (ours)
& \textbf{99.8}
& \textbf{99.6}
& \textbf{99.4}
& \textbf{95.2}
& \textbf{98.5} \\

\midrule

\multirow{4}{*}{\shortstack{Token\\number}}
& $K=16$
& 99.0 & 98.8 & 98.4 & 94.6 & 97.7 \\

& $K=36$
& 99.4 & 99.2 & 98.8 & 94.8 & 98.1 \\

& $K=64$ (ours)
& \textbf{99.8}
& \textbf{99.6}
& \textbf{99.4}
& \textbf{95.2}
& \textbf{98.5} \\

& $K=100$
& 99.6
& \textbf{99.6}
& \textbf{99.4}
& \textbf{95.2}
& \textbf{98.5} \\

\bottomrule
\end{tabular}%
}

\end{minipage}

\vspace{-4pt}
\end{table}


\section{Extended Quantitative Analyses}
\label{app:section_b}

\subsection{Extended Platform Results}
\paragraph{Full Experimental Results on RoboTwin 2.0.}
Tab.~\ref{tab:robotwin_random_results} reports the per-task results over all 50 tasks under the hard randomized setting, with systematic comparisons against ACT~\cite{zhao2023act}, Diffusion policy~\cite{diff}, 3D Diffusion policy~\cite{Ze2024DP3}, RDT-1B~\cite{liu2024rdt}, $\pi_0$~\cite{pi0}, $\pi_{0.5}$~\cite{pi05}, TwinVLA (ICLR 2026)~\cite{im2025twinvla}, BagelVLA~\cite{hu2026bagelvla}, and HALO (RA-L 2026)~\cite{shou2026halo}. 
Although VLDA is trained only on clean demonstrations and directly evaluated under OOD variations without further adaptation, it achieves the highest average success rate of \textbf{50.86\%}, outperforming the strongest baseline, $\pi_{0.5}$, by \textbf{7.02} percentage points. 
Compared with these representative imitation-learning, diffusion-based, and VLA baselines, VLDA achieves leading or tied-leading performance across a wide range of tasks involving spatial reasoning, dynamic interaction, long-horizon execution, and bimanual coordination, demonstrating robust generalization to diverse environmental variations.

\paragraph{Real-World Performance on the ALOHA Platform.}As shown in Tab.~\ref{tab:real_world_aloha}, VLDA remains the strongest method on the ALOHA platform, achieving an average complete-task success rate of \textbf{72.0\%}. This result exceeds $\pi_{0.5}$~\cite{pi05} by \textbf{18.0} percentage points and OpenVLA-OFT~\cite{oft} by \textbf{55.0} points. These results suggest that VLDA generalizes effectively across sequential assembly, dynamic object interaction, tool-mediated manipulation, and coordinated dual-arm execution.

\subsection{Extended Ablation Results}

\paragraph{Implementation Details of Transition-Paradigm and Training-Strategy Ablations.}
\textbf{1) Full-Future Prediction.}
This variant retains the same vision-language backbone, World Expert, output shape ($K=64$, $d_\Delta=256$), flow formulation, training schedule, DAMG, DSAF, and Action Expert as VLDA. The only difference lies in the supervision target. Instead of constructing current-to-future Delta and applying LT-Filter and GT-Anchor, we encode the future primary-camera observation with the EMA V-JEPA 2 teacher, adaptively pool its 256 patch tokens to 64 tokens, and project them from $d_v=1408$ to 256 dimensions. The resulting full-future tokens directly supervise the World Expert using the same $x$-prediction and entropy-regularized set-matching objectives. The predicted tokens are then passed through the unchanged World-to-Action pathway.

\begin{figure*}[t]
\centering
\includegraphics[width=1.0\textwidth]{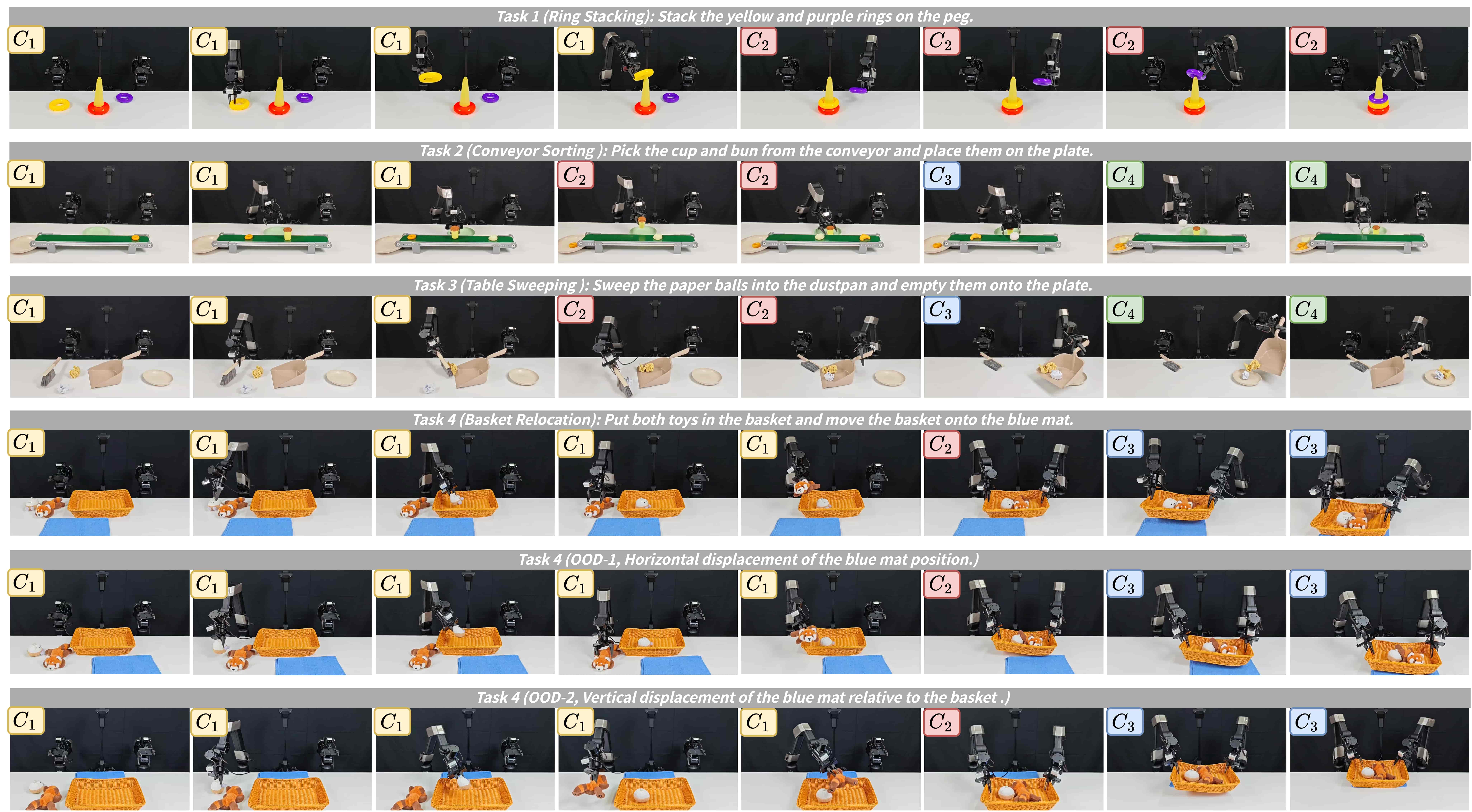}
\caption{Representative VLDA executions on the GALAXEA R1 Lite platform across four long-horizon tasks and two OOD configurations of Basket Relocation. Frames proceed chronologically from left to right, with colored labels indicating cumulative evaluation checkpoints.}
\label{fig:supp-vis1}
\end{figure*}

\begin{table*}[t]
\centering
\caption{Complete ablation results of Delta-to-Action Manifold Grounding on LIBERO and LIBERO-plus.}
\vspace{-5pt}
\label{tab:ablation_damg}
\footnotesize
\setlength{\tabcolsep}{2.8pt}
\renewcommand{\arraystretch}{1.08}
\resizebox{\textwidth}{!}{%
\begin{tabular}{l|cccc|c|ccccccc|cccc|c}
\toprule
\multirow{2}{*}{\textbf{Method}}
& \multicolumn{5}{c|}{\textbf{LIBERO}}
& \multicolumn{12}{c}{\textbf{LIBERO-plus}} \\
\cmidrule(lr){2-6}
\cmidrule(lr){7-18}
& \textbf{Spat.}
& \textbf{Obj.}
& \textbf{Goal}
& \textbf{Long}
& \textbf{Avg.}
& \textbf{Cam.}
& \textbf{Robot}
& \textbf{Lang.}
& \textbf{Light}
& \textbf{Bkg.}
& \textbf{Noise} 
& \textbf{Layout} 
& \textbf{Spat.}
& \textbf{Obj.}
& \textbf{Goal}
& \textbf{Long}
& \textbf{Avg.} \\
\midrule
w/o DAMG 
& 99.2 & 99.4 & \textbf{99.4} & 94.6 & 98.2 
& 69.5 & 74.1 & 82.5 & 89.7 & 91.8 & 86.9 & 84.6
& 86.7 & 89.1 & 76.9 & 75.5 & 82.0 \\

\textbf{VLDA}
& \textbf{99.8} & \textbf{99.6} & \textbf{99.4} & \textbf{95.2} & \textbf{98.5}
& \textbf{72.4} & \textbf{75.3}  & \textbf{84.0}  & \textbf{96.0}  & \textbf{95.5} & \textbf{89.4} & \textbf{85.7} 
& \textbf{89.6} & \textbf{89.8} & \textbf{80.1} & \textbf{78.3} & \textbf{84.5} \\
\bottomrule
\end{tabular}%
}
\end{table*}

\begin{figure*}[t]
\centering
\includegraphics[width=1.0\textwidth]{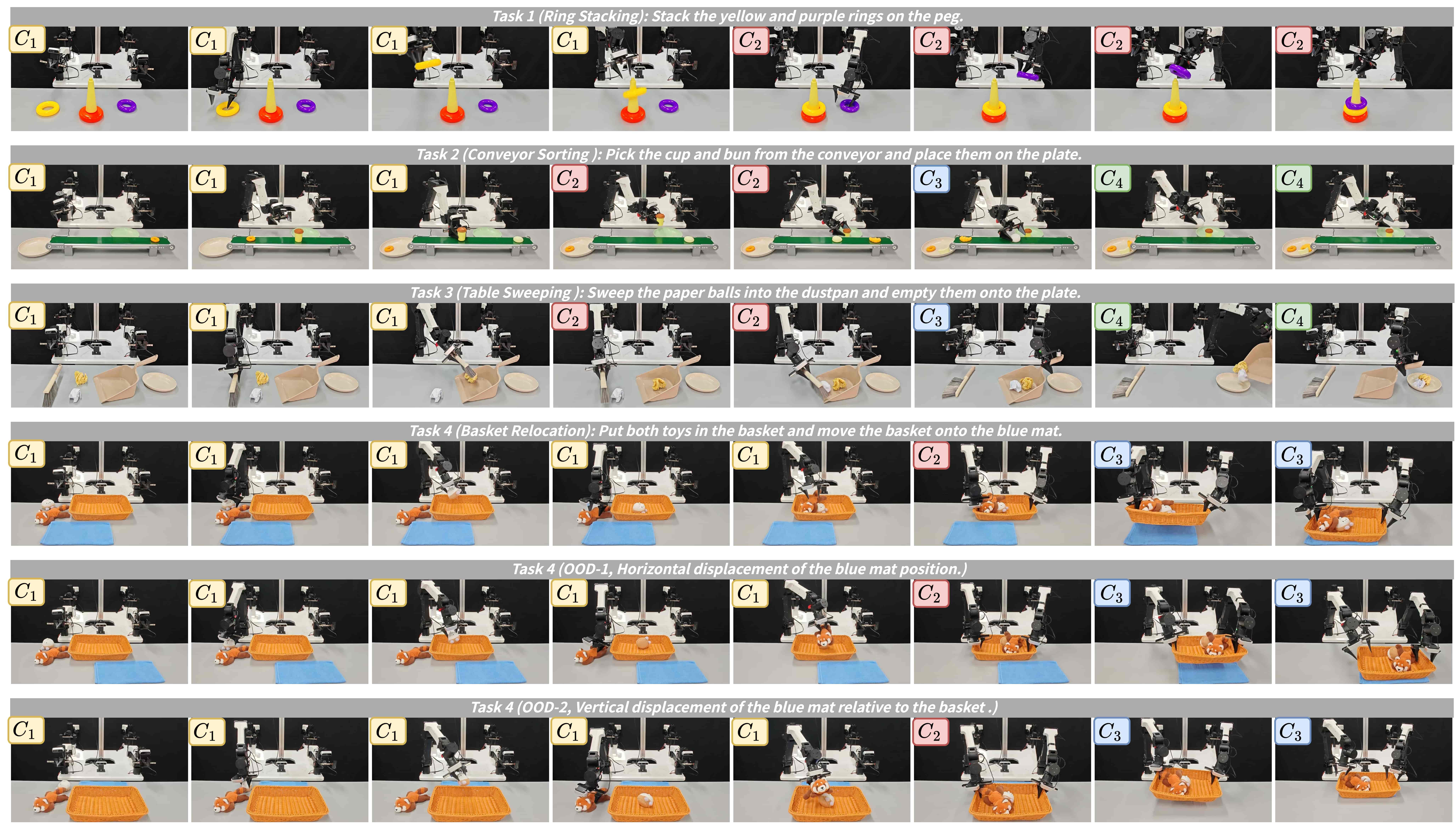}
\caption{Representative VLDA executions on the AgileX Cobot Magic platform across four long-horizon tasks and two OOD configurations of Basket Relocation. Frames proceed chronologically from left to right, with colored labels indicating cumulative evaluation checkpoints.}
\label{fig:supp-vis2}
\end{figure*}

\begin{figure*}[t]
\centering
\includegraphics[width=\textwidth]{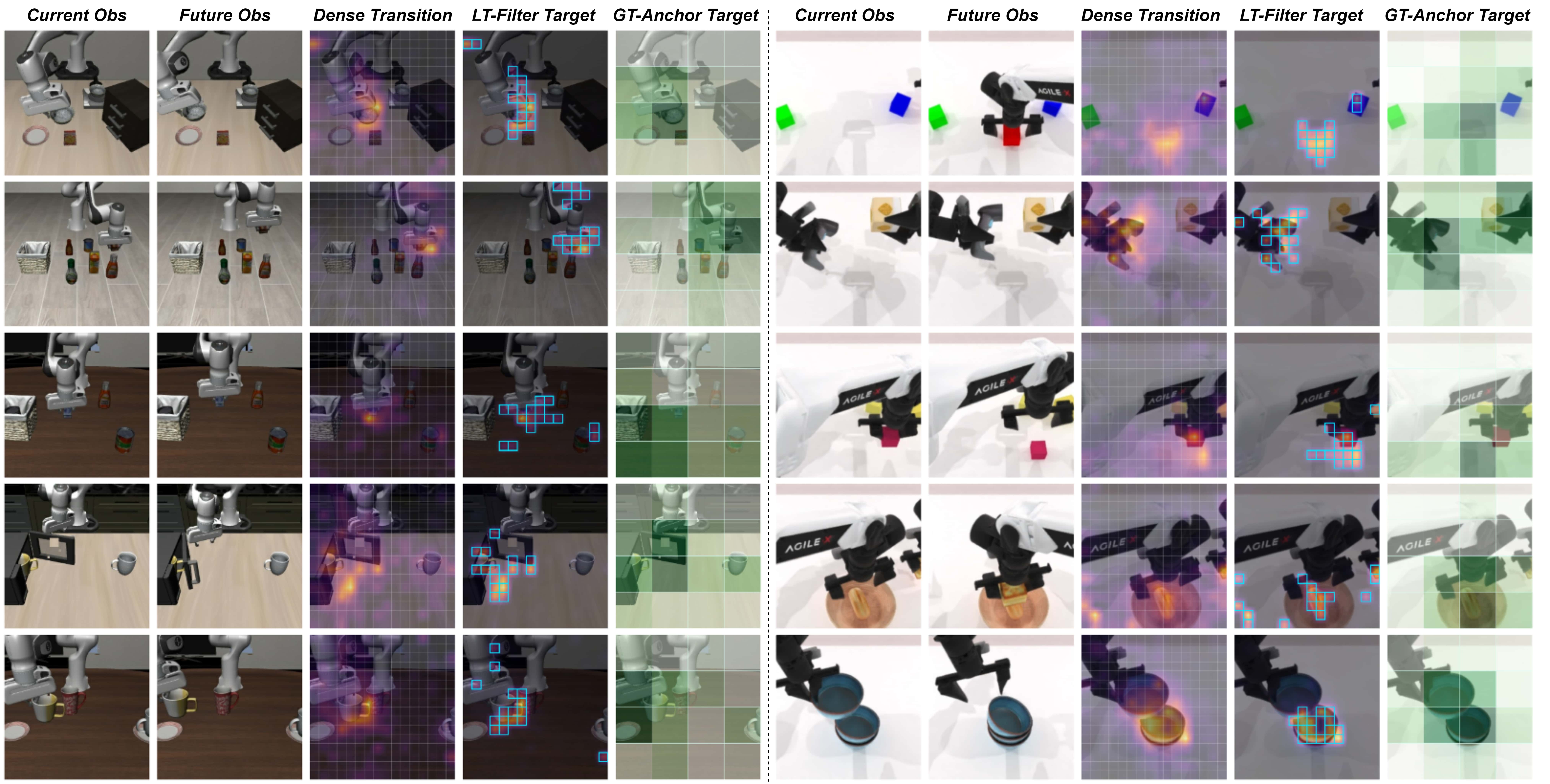}
\caption{\textbf{Visualization of Delta supervision.}
LT-Filter extracts sparse manipulation-centric local changes, while GT-Anchor preserves the coarse global transition trend.}
\label{fig:app-vis2}
\end{figure*}

\textbf{2) Without Transition Pre-Training.}
To evaluate the contribution of the two-stage training strategy, this
variant removes Stage I and directly performs joint optimization from
the same model initialization as VLDA. The vision-language backbone,
online V-JEPA 2 encoder, World Expert, Action Expert, and all projection modules are optimized jointly from the beginning using the complete second-stage objective. The EMA teacher remains gradient-free and is updated exclusively through EMA, while the Sinkhorn transport plan remains detached as in the full model. This variant is trained for the same total number of optimization steps as VLDA, with all other architectural settings, loss weights, and inference procedures kept unchanged.

As shown in Tab.~\ref{tab:libero_plus_ablation}, VLDA consistently
outperforms both variants under distribution shifts. Replacing Delta
modeling with full-future prediction reduces the overall success rate
from 84.5\% to 81.1\%, with larger drops under lighting
($-6.9$), background ($-4.7$), camera ($-4.4$), and noise
($-3.9$) perturbations. This indicates that directly modeling the
complete future representation retains more appearance-dependent and
action-irrelevant information, whereas compact Delta provides a more
selective transition interface. 

Removing transition pre-training further decreases the overall result to 79.4\%, with pronounced degradation on the Long suite ($-6.1$) and under lighting ($-8.3$), background ($-6.1$), camera ($-5.8$), and noise ($-5.1$) shifts. These results suggest that transition pre-training establishes a more stable and robust transition representation before joint action optimization.

\paragraph{Ablation of Delta-Seeded Action Flow.} 
For Delta-Seeded Action Flow, we conduct a complete ablation of its two Delta-conditioning branches: the Delta-induced noise bias and the Delta prefix token. Specifically, we evaluate three variants: removing both branches, retaining only the noise bias, and retaining only the prefix token. As shown in Tab.~\ref{tab:ablation_dsaf}, both branches contribute to action generation. Using only the Delta-induced noise bias improves the average success rate from 97.2\% to 97.9\%, while using only the Delta prefix token yields 97.6\%. Their combination achieves the best performance of 98.5\%, indicating that the noise bias provides a more informative flow initialization, while the prefix token offers complementary transition conditioning throughout denoising.

\paragraph{Ablation of Delta-to-Action Manifold Grounding.} Tab.~\ref{tab:ablation_damg} reports the complete results. Manifold Grounding is applied only during training to align Delta with the current action chunk, introducing no additional inference cost. It yields a modest 0.3-point gain on the nearly saturated LIBERO benchmark, but improves LIBERO-plus by 2.5 points on average, with larger gains under lighting (+6.3), background (+3.7), camera (+2.9), and noise (+2.5) perturbations. Since negative actions are sampled from other temporal regions of the same trajectory, they share the task and scene semantics but correspond to different execution stages. The objective therefore encourages Delta to encode transition cues specific to the current action phase, reducing its reliance on visually variable yet action-irrelevant information.

\paragraph{Ablation of Delta-Induced Noise Bias.}
Beyond the branch-wise ablation above, we further investigate how
Delta should be incorporated into action-flow initialization. To
isolate the noise-bias design, all variants retain the Delta prefix
token and differ only in how the initial action source is constructed.
We compare the standard Gaussian source without noise shifting, direct
injection of the complete Delta-token set, and the proposed mean-pooled action-space bias.

For direct injection, we preserve all Delta tokens and transform them
into the shape of the action chunk. Given
$\hat{\Delta}_t\in\mathbb{R}^{K\times d_\Delta}$, we first project
each Delta token into the action dimension:
\begin{equation}
    Q_t
    =
    q_{\mathrm{dim}}(\hat{\Delta}_t)
    \in
    \mathbb{R}^{K\times d_a},
\end{equation}
where
$q_{\mathrm{dim}}:\mathbb{R}^{d_\Delta}\rightarrow
\mathbb{R}^{d_a}$ is applied independently to each token. We then
mix the token dimension using a learnable projection
$W_{\mathrm{tok}}\in\mathbb{R}^{H\times K}$:
\begin{equation}
    \widetilde{\Delta}_t^{A}
    =
    W_{\mathrm{tok}}Q_t
    \in
    \mathbb{R}^{H\times d_a}.
\end{equation}
The transformed Delta is directly added to the Gaussian action source:
\begin{equation}
    \epsilon_A'
    =
    \epsilon_A
    +
    \gamma\widetilde{\Delta}_t^{A},
    \qquad
    \epsilon_A\sim\mathcal{N}(\mathbf{0},\mathbf{I}).
\end{equation}
By contrast, the proposed bias first aggregates Delta into the global
transition representation $\bar{\Delta}_t$ and predicts the
action-space bias $q_b(\bar{\Delta}_t)\in\mathbb{R}^{H\times d_a}$,
as defined in the main paper. For the no-shift variant, we simply use
$\epsilon_A'=\epsilon_A$, following the original action-flow
initialization of $\pi_{0.5}$.

As shown in Tab.~\ref{tab:ablation_bias}, direct Delta injection
improves the average success rate from 97.6\% to 98.0\%, confirming
that transition-informed flow initialization facilitates action
denoising. The proposed bias further improves performance to 98.5\%.
Since Delta is learned as an unordered token set, directly projecting
all tokens into ordered action positions introduces an additional
token-mixing burden. Mean pooling instead provides a
permutation-invariant transition summary, which can be adaptively
mapped into a more stable and effective action-space bias.


\paragraph{Ablation of Transition-Token Selection.}
We examine whether LT-Filter benefits from selecting salient transition regions rather than merely reducing the token number. We compare uniformly sampled \emph{Random-$K$}, adaptive \emph{Spatial pooling}, and the proposed \emph{Top-$K$} based on $\ell_2$ transition magnitudes. As shown in Tab.~\ref{tab:ablation_for_3}, Top-$K$ outperforms Random-$K$ and spatial pooling by 1.1 and 0.6 points, respectively. This confirms that explicitly retaining regions with salient changes is more effective than random selection or spatial compression.

\paragraph{Ablation of Set-Matching Objectives.}
We compare three objectives for supervising unordered Delta tokens. Point-wise MSE assumes index-wise correspondence, Chamfer distance uses bidirectional nearest-neighbor matching, and entropy-regularized OT performs soft balanced matching. Point-wise MSE and Chamfer distance achieve average success rates of 97.2\% and 97.8\%, respectively, while OT reaches 98.5\%. This demonstrates that soft balanced matching provides more effective supervision than fixed correspondence or independent nearest-neighbor assignments.

\paragraph{Ablation of the Number of Delta Tokens.}
We evaluate $K\in\{16,36,64,100\}$ while retaining Top-$K$ selection and entropy-regularized OT. Increasing $K$ from 16 to 36 and 64 improves the average success rate from 97.7\% to 98.1\% and 98.5\%, respectively, as more relevant local transitions are preserved. Increasing $K$ to 100 provides no further average improvement while incurring additional computation. We therefore use $K=64$ as the
default, balancing transition coverage and computational efficiency.






\section{Extended Qualitative Analyses}
\label{app:section_c}

\subsection{Extended Real-World Task Execution Visualization}

\paragraph{Execution Visualizations on the GALAXEA R1 Lite Platform.} Fig.~\ref{fig:supp-vis1} presents representative execution trajectories of VLDA on the four long-horizon real-world tasks, together with two OOD spatial configurations of Basket Relocation. Frames are arranged chronologically from left to right, and the colored labels indicate the cumulative evaluation checkpoints defined in Tab.~\ref{tab:real_task_protocol}. These visualizations demonstrate that VLDA maintains consistent task progress across sequential interaction stages rather than relying on isolated observation-to-action mappings.

\begin{itemize}
    
    \item For \textit{Ring Stacking}, VLDA first identifies and places the yellow ring around the peg, and subsequently stacks the purple ring above it, successfully preserving the semantic order specified by the instruction.

    \item In \textit{Conveyor Sorting}, the policy intercepts the moving cup and bun in sequence, places the cup upright on the plate, and avoids selecting distractor objects. This task requires both dynamic target tracking and instruction-conditioned object selection.

    \item For \textit{Table Sweeping}, VLDA progressively sweeps the two paper balls into the dustpan, lifts the dustpan while preserving its contents, and finally empties both balls onto the plate. The successful execution illustrates robust contact-rich tool use and consistent maintenance of intermediate task states.

    \item \textit{Basket Relocation} further evaluates long-horizon bimanual coordination and spatial generalization. VLDA sequentially places both toys into the basket, establishes a stable bimanual grasp, and relocates the basket onto the blue mat without dropping either object. Beyond the standard configuration, the bottom two rows visualize OOD settings in which the blue mat is displaced horizontally or vertically relative to the basket. Despite these spatial shifts, VLDA adapts the transport trajectory and completes all checkpoints, indicating that the learned Delta captures the task-relevant transition toward the target region rather than overfitting to fixed absolute layouts. 
    
\end{itemize}

\paragraph{Execution Visualizations on the AgileX Cobot Magic Platform.} Fig.~\ref{fig:supp-vis2} presents representative VLDA execution trajectories for the four long-horizon tasks on the AgileX Cobot Magic platform, including two OOD spatial configurations of \textit{Basket Relocation}. Frames proceed chronologically from left to right, with colored labels denoting the cumulative checkpoints in Tab.~\ref{tab:real_task_protocol}. Across sequential ring stacking, dynamic conveyor sorting, contact-rich table sweeping, and coordinated basket relocation, VLDA consistently completes the required interaction stages while preserving task order and intermediate states. Its successful adaptation to horizontal and vertical displacements of the blue mat further demonstrates robust spatial generalization beyond the training layouts.

\subsection{Extended Delta Supervision Visualization}

Fig.~\ref{fig:app-vis2} provides additional visualizations of the Delta supervision on LIBERO (left) and RoboTwin 2.0 (right). For each example, the five columns show the current observation, future observation, dense transition, LT-Filter target, and GT-Anchor target, respectively. The dense transition is computed from the feature difference between the future and current observations, with higher-intensity regions indicating stronger temporal changes.

LT-Filter converts these dense differences into sparse, manipulation-centric targets. Across both benchmarks, the selected patches consistently concentrate around task-relevant objects, end-effectors, contact points, and regions undergoing physical state changes, while suppressing static backgrounds and unrelated scene content. This behavior indicates that LT-Filter isolates the local transition evidence most directly associated with the action being executed. GT-Anchor aggregates future information into a coarse scene-level representation. Its lower-resolution targets preserve the overall spatial direction and structural evolution of the scene, complementing the fine-grained selectivity of LT-Filter. LT-Filter and GT-Anchor enable Delta to retain sparse action-relevant changes while remaining globally aligned with future state evolution, thereby providing a compact and robust transition representation for subsequent action generation.

\section{Discussion}
\label{app:section_d}

\subsection{Collapse Mitigation in Transition Representation Learning}
\label{app:collapse_prevention}

In this section, we explain the asymmetric encoder design used for
transition representation learning. VLDA employs an online encoder and
an EMA teacher encoder to stabilize transition supervision and reduce
the risk of representation collapse. Let $E_{\psi}$ and
$E_{\bar{\psi}}$ denote the online and EMA teacher encoders,
respectively. The online encoder is optimized by gradients, whereas
the EMA teacher is updated exclusively through exponential moving
average:
\begin{equation}
    \bar{\psi}_{r}
    =
    m\bar{\psi}_{r-1}
    +
    (1-m)\psi_r,
    \qquad
    m\in[0,1).
\end{equation}
This design follows the asymmetric optimization principle of
V-JEPA 2~\cite{vjepa2} and provides a slowly evolving reference for
current-to-future transition learning.

\paragraph{Asymmetric Gradient Flow.}
The current and future primary-view representations are computed as
\begin{equation}
    \mathbf V_t
    =
    E_{\psi}(I_t^{\mathrm{pri}}),
    \qquad
    \bar{\mathbf V}_{t+H}
    =
    \operatorname{sg}
    \left(
    E_{\bar{\psi}}(I_{t+H}^{\mathrm{pri}})
    \right),
\end{equation}
where $\operatorname{sg}(\cdot)$ denotes stop-gradient. Their dense
transition is
\begin{equation}
    \mathbf D_t
    =
    \bar{\mathbf V}_{t+H}
    -
    \mathbf V_t.
\end{equation}
The compact Delta target is then constructed as
\begin{equation}
\begin{aligned}
    \mathcal K_t
    &=
    \operatorname{TopK}_{K}
    \left(
    \{\|\mathbf D_{t,i}\|_2\}_{i=1}^{P}
    \right),\\
    \Delta_t^\star
    &=
    g_{\downarrow}
    \left(
    \mathbf D_{t,\mathcal K_t}
    \right).
\end{aligned}
\end{equation}

Gradients propagate through the selected transition features and the
down-projection module, while the discrete Top-$K$ indices are treated
as fixed within each backward pass. No gradient is propagated through
the future target branch:
\begin{equation}
    \frac{\partial\Delta_t^\star}{\partial\bar{\psi}}
    =
    \mathbf 0,
    \qquad
    \frac{\partial\Delta_t^\star}{\partial\psi}
    \neq
    \mathbf 0,
    \qquad
    \frac{\partial\Delta_t^\star}
    {\partial\omega_{\downarrow}}
    \neq
    \mathbf 0,
\end{equation}
where $\omega_{\downarrow}$ denotes the parameters of
$g_{\downarrow}$. This asymmetric gradient flow keeps the future
endpoint fixed within each optimization step and prevents direct
gradient-based co-adaptation between the online and teacher branches.

\paragraph{Slowly Evolving Target.}
The EMA teacher evolves substantially more slowly than the online
encoder. From the EMA update,
\begin{equation}
    \bar{\psi}_{r}
    -
    \bar{\psi}_{r-1}
    =
    (1-m)
    \left(
    \psi_r-\bar{\psi}_{r-1}
    \right),
\end{equation}
and consequently
\begin{equation}
    \left\|
    \bar{\psi}_{r}
    -
    \bar{\psi}_{r-1}
    \right\|
    =
    (1-m)
    \left\|
    \psi_r-\bar{\psi}_{r-1}
    \right\|.
\end{equation}
When $m$ is close to one, only a small fraction of the discrepancy
between the online and teacher parameters is incorporated at each
iteration.

Assuming that the encoder is locally $L_E$-Lipschitz with respect to
its parameters, the variation of the future representation satisfies
\begin{equation}
\begin{aligned}
    &
    \left\|
    E_{\bar{\psi}_r}(I_{t+H}^{\mathrm{pri}})
    -
    E_{\bar{\psi}_{r-1}}(I_{t+H}^{\mathrm{pri}})
    \right\|\\
    &\qquad
    \le
    L_E(1-m)
    \left\|
    \psi_r-\bar{\psi}_{r-1}
    \right\|.
\end{aligned}
\end{equation}
The future endpoint therefore evolves smoothly rather than changing
synchronously with the online representation.

\paragraph{Complementary Transition Constraints.}
During Stage I, GT-Anchor requires the predicted compact Delta to
explain the pooled transition in the high-dimensional visual feature
space. Under the pretrained and non-degenerate visual feature space,
these pooled transition anchors vary across samples and execution
stages, thereby discouraging the predicted Delta from collapsing to a
sample-independent constant representation. GT-Anchor therefore
encourages Delta to preserve sample-dependent transition information.

During Stage II, Delta-to-Action Manifold Grounding further requires
Delta to distinguish actions in the current chunk from actions sampled
from other temporal regions of the same trajectory. Since these
negative actions share the same task, scene, and embodiment context but
correspond to different execution stages, the alignment objective
encourages Delta to retain phase-specific and action-relevant
transition information.

The action-generation loss also propagates through the Delta-induced
noise bias and Delta prefix token, providing additional
action-dependent supervision to the World Expert and the predicted
Delta representation.

\paragraph{Transition Representation Stabilization.}
Pretrained initialization, asymmetric gradient flow, and slow EMA
updates provide a stable and structured future reference for
current-to-future transition learning. GT-Anchor encourages
sample-dependent global transition information, while
Delta-to-Action Manifold Grounding and the action-generation objective
further constrain Delta to encode phase-specific and
execution-relevant factors. Together, these complementary visual,
temporal, and action-grounded constraints reduce rapid co-adaptation
and help maintain an informative and actionable Delta space throughout
optimization.

\subsection{Entropy-Regularized Transport Matching}
\label{app:set_distance}

The token collections compared in LT-Filter and GT-Anchor are treated
as unordered sets. Specifically, LT-Filter matches the predicted and
target Delta-velocity tokens, while GT-Anchor matches the projected
Delta tokens with the global transition anchors. Point-wise regression
would impose an arbitrary index-wise correspondence between these
tokens. We therefore employ entropy-regularized optimal transport to
construct a permutation-invariant soft matching between the two sets.

Let
\begin{equation}
    \mathbf X
    =
    [\mathbf x_1,\ldots,\mathbf x_K]^\top,
    ~~
    \mathbf Y
    =
    [\mathbf y_1,\ldots,\mathbf y_K]^\top
    \in
    \mathbb R^{K\times d}.
\end{equation}
We assign uniform mass
$\mathbf a=\mathbf b=\frac{1}{K}\mathbf 1_K$ and define the balanced
transport polytope as
\begin{equation}
    \Pi(\mathbf a,\mathbf b)
    =
    \left\{
    \mathbf T\in\mathbb R_+^{K\times K}
    \ \middle|\
    \mathbf T\mathbf 1_K=\mathbf a,\,
    \mathbf T^\top\mathbf 1_K=\mathbf b
    \right\}.
\label{eq:transport_polytope}
\end{equation}
These constraints ensure that every token in both sets contributes the
same total transport mass.

The pairwise token cost is defined as the feature-normalized squared
Euclidean distance:
\begin{equation}
    C_{ij}
    =
    \frac{
    \|\mathbf x_i-\mathbf y_j\|_2^2
    }{d}.
\label{eq:ot_cost}
\end{equation}
The dimensional normalization enables the same set-matching objective
to be applied in both the compact Delta space of dimension
$d_\Delta$ and the visual feature space of dimension $d_v$.

To improve numerical stability when estimating the transport plan, we
normalize the cost matrix independently for each sample:
\begin{equation}
    s
    =
    \max \left( \operatorname{sg} \left[
    \frac{1}{K^2} \sum_{i,j}C_{ij}
    \right], 10^{-6} \right),
    ~~
    \bar C_{ij} =
    \frac{C_{ij}}{s},
\label{eq:ot_plan_cost}
\end{equation}
where $\operatorname{sg}(\cdot)$ denotes stop-gradient. The normalized
cost $\bar{\mathbf C}$ is used only to estimate the transport plan,
whereas the final set-matching loss is evaluated using the original
cost matrix $\mathbf C$.

The entropy-regularized transport plan is defined as
\begin{equation}
    \mathbf T_\eta^\star
    =
    \arg\min_{\mathbf T\in\Pi(\mathbf a,\mathbf b)}
    \left[
    \langle\mathbf T,\bar{\mathbf C}\rangle
    +
    \eta
    \sum_{i,j}
    T_{ij}\log T_{ij}
    \right],
\label{eq:ot_plan}
\end{equation}
where
$\langle\mathbf T,\bar{\mathbf C}\rangle =\sum_{i,j}T_{ij}\bar C_{ij}$, $\eta>0$ controls the softness of the correspondence, and
$0\log 0=0$. A smaller value of $\eta$ produces a sharper
correspondence, whereas a larger value distributes the transport mass
over more token pairs.

We solve Equation~\eqref{eq:ot_plan} using log-domain Sinkhorn
iterations. Let
\begin{equation}
    L_{ij}
    =
    -\frac{\bar C_{ij}}{\eta}.
\end{equation}
Starting from zero log-scaling vectors $\mathbf p$ and $\mathbf q$, we
alternately update
\begin{equation}
\begin{aligned}
    p_i
    &\leftarrow
    \log a_i
    -
    \operatorname{LSE}_{j}
    \left(
    L_{ij}+q_j
    \right),\\
    q_j
    &\leftarrow
    \log b_j
    -
    \operatorname{LSE}_{i}
    \left(
    L_{ij}+p_i
    \right),
\end{aligned}
\label{eq:sinkhorn_updates}
\end{equation}
where $\operatorname{LSE}$ denotes the log-sum-exp operation. The
resulting transport plan is
\begin{equation}
    T_{\eta,ij}^\star
    =
    \exp
    \left(
    L_{ij}+p_i+q_j
    \right).
\end{equation}
We use 50 Sinkhorn iterations and set $\eta=0.05$ in all experiments.

The transport plan is used only to determine the soft correspondence.
We therefore stop gradients through the Sinkhorn solver:
\begin{equation}
    \widetilde{\mathbf T}_\eta
    =
    \operatorname{sg}
    \left(
    \mathbf T_\eta^\star
    \right).
\end{equation}
The set-matching loss used in the main paper is implemented as
\begin{equation}
    \mathcal D_{\mathrm{set}}
    (\mathbf X,\mathbf Y;d)
    =
    \left\langle
    \widetilde{\mathbf T}_\eta,
    \mathbf C
    \right\rangle
    =
    \sum_{i,j}
    \widetilde T_{\eta,ij}C_{ij}.
\label{eq:implemented_set_loss}
\end{equation}
Thus, entropy regularization influences optimization through the soft
correspondence encoded by $\widetilde{\mathbf T}_\eta$, while the
entropy term itself is not added to the final scalar training loss.
Gradients propagate through the original pairwise costs to the token
features without back-propagating through the Sinkhorn iterations.

Finally, let $\mathbf P$ and $\mathbf R$ be arbitrary permutation
matrices applied independently to the two token sets. The corresponding
cost and transport matrices are permuted as
$\mathbf P\mathbf C\mathbf R^\top$ and
$\mathbf P\mathbf T_\eta^\star\mathbf R^\top$, respectively.
Consequently,
\begin{equation}
    \mathcal D_{\mathrm{set}}
    (\mathbf P\mathbf X,\mathbf R\mathbf Y;d)
    =
    \mathcal D_{\mathrm{set}}
    (\mathbf X,\mathbf Y;d).
\end{equation}
The objective therefore requires neither fixed token identities nor a
predefined ordering of the Delta or transition-anchor tokens.

\subsection{Supplementary Details on Motivation}
\label{app:motivation}

This section further explains the design considerations behind
actionable Delta modeling, including the choice of transition targets,
the complementary supervision objectives, and the role of Delta in
action generation.

\paragraph{Why an Actionable Transition Interface?}
Direct observation-to-action policies must implicitly infer the
desired physical change and translate it into motor commands within a
single representation. Complete future-state prediction introduces
temporal structure, but its target contains both the state information
already present in the current observation and the comparatively sparse
changes caused by manipulation. Consequently, a large portion of its
learning capacity may be devoted to reconstructing static objects,
background appearance, and other information that does not determine
the next action.

Delta provides an intermediate alternative between these two formulations. Rather than reproducing the complete future state or mapping semantic observations directly to actions, it explicitly represents the transition between the current and future states. Feature differencing suppresses much of the shared appearance content, while the compact token space further limits residual redundancy. Delta is therefore not intended to reconstruct every future detail; it serves as a control-oriented interface that conveys the physical transition most useful for subsequent action generation.

\paragraph{Why Horizon-Aligned Future Supervision?}
The future observation at $t+H$ is aligned with the action chunk
$\mathbf{A}_t=[a_t,\ldots,a_{t+H-1}]$. It records the state reached
after executing the actions covered by the current prediction horizon,
making the current-to-future difference a natural supervision signal
for the physical effect of that chunk. A substantially shorter
transition would describe only an early fraction of the predicted
motion, whereas a substantially longer transition could include state
changes beyond the temporal scope of the current action chunk.

\paragraph{Why Combine Local and Global Transition Supervision?}
A useful transition representation must be selective without becoming
myopic. LT-Filter addresses selectivity by retaining regions with large
feature-transition magnitudes. This suppresses the many spatial tokens
whose current and future representations remain nearly unchanged and
emphasizes candidate interaction regions. However, magnitude-based
selection alone can overemphasize isolated local changes or discard
distributed transitions, such as the overall displacement of a large
object.

GT-Anchor provides the complementary constraint. By pooling before
computing the transition, it retains a coarse summary of the overall
state evolution without requiring reconstruction of the complete
future feature map. LT-Filter can therefore be viewed as preserving
high-resolution local evidence, while GT-Anchor preserves low-resolution
global consistency. Their combination encourages Delta to capture
manipulation-centric details while remaining compatible with the
broader direction of future evolution.

\paragraph{Why Ground Delta in the Action Manifold?}
Visual transition supervision alone does not guarantee action
relevance. Feature differences may also reflect illumination changes,
background motion, or object changes that are visually salient but
unrelated to the demonstrated behavior. Delta-to-Action Manifold
Grounding addresses this ambiguity by comparing every action in the
current chunk with the predicted Delta-token set. Actions from the
corresponding chunk serve as positives, while actions from
non-overlapping regions of the same trajectory serve as negatives.

Sampling negatives from the same trajectory is important because it
controls for much of the shared scene, task, and embodiment context.
The alignment objective must therefore distinguish the action
associated with the current local transition from other individually
plausible actions in the same demonstration. As detailed in
Appendix~\ref{app:math_derivations}, this action-wise InfoNCE objective
admits a conditional mutual-information interpretation. It encourages
Delta to retain transition factors that are informative about all
actions in the current chunk rather than merely encoding visually
predictive appearance changes.

\paragraph{Why Two-Stage Optimization?}
The World Expert and the Delta pathway are newly introduced components,
whereas the VLM and Action Expert inherit strong pretrained
representations. If all objectives are optimized jointly from the
beginning, the pretrained action pathway can reduce the action loss
without consistently relying on the initially unstructured Delta
representation. This can weaken the role of Delta as an explicit
transition interface.

The first stage therefore optimizes LT-Filter and GT-Anchor before
joint action training, allowing the World Expert to establish a stable
and future-grounded transition space. The second stage then introduces
action generation and Delta--Action alignment, grounding this learned
transition space in executable behavior. The observed degradation
under one-stage training further supports the benefit of establishing the transition interface before joint action refinement.

\subsection{Mathematical Details of Key Equations}
\label{app:math_derivations}

In this section, we provide mathematical details for the key objectives in VLDA, including the expert-specific flow parameterizations and the mutual information maximization perspective of Delta-to-Action Manifold Grounding.

\paragraph{Expert-Specific Flow Parameterizations.}
The Action and World Experts share the same linear conditional path
but use different source distributions. For $E\in\{A,\Delta\}$, let
\begin{equation}
\begin{aligned}
    (\mathbf{x}_0^A,\mathbf{z}_A)
    &=
    (\mathbf{A}_t,\epsilon_A+\gamma\mathbf{b}_t),\\
    (\mathbf{x}_0^\Delta,\mathbf{z}_\Delta)
    &=
    (\Delta_t^\star,\epsilon_\Delta).
\end{aligned}
\end{equation}
Conditioned on the predicted action-space bias $\mathbf b_t$, the
corresponding source distributions are
\begin{equation}
    \mathbf{z}_A
    \sim
    \mathcal{N}(\gamma\mathbf{b}_t,\mathbf I),
    \qquad
    \mathbf{z}_\Delta
    \sim
    \mathcal{N}(\mathbf 0,\mathbf I).
\end{equation}
During Stage II, $\mathbf b_t$ remains differentiable, allowing the
action-generation objective to update $\gamma$, $q_b$, and the
upstream Delta pathway.

Both experts follow the linear conditional path
\begin{equation}
    \mathbf{x}_\tau^E
    =
    (1-\tau)\mathbf{x}_0^E+\tau\mathbf{z}_E,
\end{equation}
which satisfies
$\mathbf{x}_{\tau=0}^E=\mathbf{x}_0^E$ and
$\mathbf{x}_{\tau=1}^E=\mathbf{z}_E$. Its derivative gives the
constant conditional velocity
\begin{equation}
    \frac{\mathrm{d}\mathbf{x}_\tau^E}{\mathrm{d}\tau}
    =
    \mathbf{z}_E-\mathbf{x}_0^E
    =
    \mathbf{u}_\tau^E.
\end{equation}
The Action Expert directly predicts this velocity. For the World
Expert, the Delta path gives, for $\tau>0$,
\begin{equation}
    \mathbf{u}_\tau^\Delta
    =
    \frac{
    \Delta_\tau-\Delta_t^\star
    }{\tau}.
\end{equation}
Replacing the clean endpoint with the World-Expert prediction gives
\begin{equation}
    \hat{\mathbf{u}}_\tau^\Delta
    =
    \frac{
    \Delta_\tau-\hat{\Delta}_t
    }{\tau},
    \qquad
    \hat{\mathbf{u}}_\tau^\Delta
    -
    \mathbf{u}_\tau^\Delta
    =
    \frac{
    \Delta_t^\star-\hat{\Delta}_t
    }{\tau}.
\end{equation}

For numerical stability, the World-Expert flow time is sampled as
\begin{equation}
    \tau
    =
    0.999\xi,
    \qquad
    \xi
    \sim
    \operatorname{Beta}(1.5,1.0).
\end{equation}
When converting $x$-predictions into velocities, we use
\begin{equation}
    \widetilde{\tau}
    =
    \max(\tau,\tau_{\min}),
    \qquad
    \tau_{\min}=0.05,
\end{equation}
and compute
\begin{equation}
    \mathbf{u}_{\tau,\mathrm{stab}}^\Delta
    =
    \frac{
    \Delta_\tau-\Delta_t^\star
    }{\widetilde{\tau}},
    \qquad
    \hat{\mathbf{u}}_{\tau,\mathrm{stab}}^\Delta
    =
    \frac{
    \Delta_\tau-\hat{\Delta}_t
    }{\widetilde{\tau}}.
\end{equation}
The same stabilized denominator is used for the target and predicted
velocities, and no additional time-dependent loss weighting is
applied. This bounds endpoint-error amplification by
$1/\tau_{\min}=20$. Therefore, the shifted action source modifies the
source boundary while preserving the linear conditional path, whereas
the World Expert uses a numerically stabilized conversion from
$x$-prediction to velocity prediction.

\paragraph{Mathematical Properties of Delta-Action Alignment.}
We further analyze the Delta-action alignment objective introduced in
Section~4.5 of the main paper. Let
$\Delta=\{\Delta_k\}_{k=1}^{K}$ denote the predicted Delta-token set,
and let $a$ denote an individual action token. For notational
simplicity, $H$ below denotes the number of positive actions used by
the alignment objective; in implementation, this number is capped at
10.

The projected and normalized representations are
\begin{equation}
\begin{aligned}
    \mathbf z_k^\Delta
    &=
    \operatorname{Norm}
    \left(
    q_\Delta(\Delta_k)
    \right),\\
    \mathbf z^{\mathrm{act}}(a)
    &=
    \operatorname{Norm}
    \left(
    q_{\mathrm{act}}(a)
    \right).
\end{aligned}
\end{equation}
Let
\begin{equation}
    c_k(\Delta,a)
    =
    \left\langle
    \mathbf z_k^\Delta,
    \mathbf z^{\mathrm{act}}(a)
    \right\rangle
\end{equation}
denote the similarity between the $k$-th Delta token and the action
token. The soft token-pooled similarity is
\begin{equation}
    s(\Delta,a)
    =
    \tau_\Delta
    \log
    \sum_{k=1}^{K}
    \exp
    \left(
    \frac{c_k(\Delta,a)}{\tau_\Delta}
    \right).
\label{eq:soft_delta_pooling}
\end{equation}

\paragraph{Permutation Invariance and Soft Token Assignment.}
Because Equation~\eqref{eq:soft_delta_pooling} sums over the complete
Delta-token set, its value is invariant to any permutation of the
Delta tokens. For an arbitrary permutation matrix $\mathbf P$,
\begin{equation}
    s(\mathbf P\Delta,a)
    =
    s(\Delta,a).
\end{equation}
Therefore, Delta-action alignment does not require a fixed token
identity or predefined token ordering.

Define the token-assignment weights as
\begin{equation}
    \alpha_k(\Delta,a)
    =
    \frac{
    \exp\left(c_k(\Delta,a)/\tau_\Delta\right)
    }{
    \sum_{j=1}^{K}
    \exp\left(c_j(\Delta,a)/\tau_\Delta\right)
    }.
\label{eq:delta_action_assignment}
\end{equation}
The derivative of the pooled similarity with respect to each
token-action similarity is
\begin{equation}
    \frac{
    \partial s(\Delta,a)
    }{
    \partial c_k(\Delta,a)
    }
    =
    \alpha_k(\Delta,a).
\end{equation}
Thus, the alignment gradient is distributed across Delta tokens
according to their compatibility with the action. Highly relevant
tokens receive larger gradients, while less relevant tokens contribute
less, without requiring a discrete token assignment.

The pooled similarity also satisfies
\begin{equation}
\begin{aligned}
    \max_k c_k(\Delta,a)
    \leq
    s(\Delta,a)
    \leq
    \max_k c_k(\Delta,a)
    +
    \tau_\Delta\log K.
\end{aligned}
\label{eq:soft_pooling_bound}
\end{equation}
As $\tau_\Delta\rightarrow0$, the pooled similarity approaches the
maximum token-action similarity:
\begin{equation}
    \lim_{\tau_\Delta\rightarrow0}
    s(\Delta,a)
    =
    \max_k c_k(\Delta,a).
\end{equation}
Consequently, soft token pooling can be interpreted as a smooth
matching between each action token and the most compatible transition
tokens. Unlike mean pooling, it preserves token-level selectivity while
remaining differentiable and permutation invariant.

\paragraph{Action-Wise Contrastive Optimization.}
The contrastive critic is defined as
\begin{equation}
    f(\Delta,a)
    =
    \frac{s(\Delta,a)}{\tau_c}.
\end{equation}
For each positive action $a_{t+h}$ and $N$ negative actions
$\{a_n^-\}_{n=1}^{N}$, the action-wise contrastive loss is
\begin{equation}
    \ell_h
    =
    -
    \log
    \frac{
    \exp f(\Delta,a_{t+h})
    }{
    \exp f(\Delta,a_{t+h})
    +
    \sum_{n=1}^{N}
    \exp f(\Delta,a_n^-)
    }.
\label{eq:action_nce_loss}
\end{equation}

Let the positive candidate have index $0$, and let
$a_0=a_{t+h}$ and $a_n=a_n^-$ for $n\geq1$. Define
\begin{equation}
    \pi_j
    =
    \frac{
    \exp f(\Delta,a_j)
    }{
    \sum_{m=0}^{N}
    \exp f(\Delta,a_m)
    }.
\end{equation}
The gradients with respect to the candidate scores are
\begin{equation}
    \frac{\partial\ell_h}{\partial f(\Delta,a_j)}
    =
    \pi_j-\mathbf{1}[j=0].
\label{eq:nce_score_gradient}
\end{equation}
In particular,
\begin{equation}
    \frac{\partial\ell_h}
    {\partial f(\Delta,a_{t+h})}
    =
    \pi_0-1,
    \qquad
    \frac{\partial\ell_h}
    {\partial f(\Delta,a_n^-)}
    =
    \pi_n.
\end{equation}
The objective therefore increases the compatibility between Delta and
the current positive action, while decreasing its compatibility with
actions from other execution stages. Negatives receiving larger
softmax probabilities also receive larger repulsive gradients,
providing an implicit emphasis on more confusable action phases.

The complete alignment objective averages over the selected actions
in the current chunk:
\begin{equation}
    \mathcal L_{\mathrm{DDAR}}
    =
    \frac{1}{H}
    \sum_{h=0}^{H-1}
    \mathbb E[\ell_h].
\end{equation}

\paragraph{Idealized Conditional InfoNCE Interpretation.}
We next provide an idealized mutual-information interpretation. Let
$\mathcal C$ denote the task, scene, and embodiment context shared by
the positive and negative candidates, excluding their temporal action
identity. Suppose that, conditioned on $\mathcal C$, the positive
action is sampled from
\begin{equation}
    a^+
    \sim
    p(a\mid\Delta,\mathcal C),
\end{equation}
whereas the negative actions are independently sampled from
\begin{equation}
    a_n^-
    \overset{\mathrm{i.i.d.}}{\sim}
    p(a\mid\mathcal C).
\end{equation}
The corresponding conditional density ratio is
\begin{equation}
    \rho_{\mathcal C}(\Delta,a)
    =
    \frac{
    p(a\mid\Delta,\mathcal C)
    }{
    p(a\mid\mathcal C)
    }.
\end{equation}

Consider a candidate set containing one positive and $N$ negative
actions, and let $J\in\{0,\ldots,N\}$ indicate the index of the
positive candidate. Under the idealized sampling process, its posterior
is
\begin{equation}
    p
    \left(
    J=j
    \mid
    \Delta,a_{0:N},\mathcal C
    \right)
    =
    \frac{
    \rho_{\mathcal C}(\Delta,a_j)
    }{
    \sum_{m=0}^{N}
    \rho_{\mathcal C}(\Delta,a_m)
    }.
\end{equation}
Accordingly, the optimal contrastive critic has the density-ratio form
\begin{equation}
    f_{\mathcal C}^\star(\Delta,a)
    =
    \log
    \frac{
    p(a\mid\Delta,\mathcal C)
    }{
    p(a\mid\mathcal C)
    }
    +
    c(\Delta,\mathcal C),
\end{equation}
where $c(\Delta,\mathcal C)$ does not depend on the action candidate
and therefore cancels in the softmax.

Under these assumptions, the standard InfoNCE result
implies~\cite{oord2019representation}
\begin{equation}
    I
    \left(
    \Delta;
    a_{t+h}
    \mid
    \mathcal C
    \right)
    \geq
    \log(N+1)
    -
    \mathcal L_h,
\label{eq:conditional_nce_bound}
\end{equation}
where $\mathcal L_h=\mathbb E[\ell_h]$. Averaging over the selected
actions gives
\begin{equation}
    \frac{1}{H}
    \sum_{h=0}^{H-1}
    I
    \left(
    \Delta;
    a_{t+h}
    \mid
    \mathcal C
    \right)
    \geq
    \log(N+1)
    -
    \mathcal L_{\mathrm{DDAR}}.
\label{eq:averaged_nce_bound}
\end{equation}

Since each $a_{t+h}$ is a component of the complete action chunk
$\mathbf A_t$, the data-processing property gives
\begin{equation}
    I
    \left(
    \Delta;
    \mathbf A_t
    \mid
    \mathcal C
    \right)
    \geq
    I
    \left(
    \Delta;
    a_{t+h}
    \mid
    \mathcal C
    \right)
\end{equation}
for every $h$. Therefore,
\begin{equation}
\begin{aligned}
    I
    \left(
    \Delta;
    \mathbf A_t
    \mid
    \mathcal C
    \right)
    &\geq
    \max_{0\leq h<H}
    I
    \left(
    \Delta;
    a_{t+h}
    \mid
    \mathcal C
    \right)\\
    &\geq
    \frac{1}{H}
    \sum_{h=0}^{H-1}
    I
    \left(
    \Delta;
    a_{t+h}
    \mid
    \mathcal C
    \right)\\
    &\geq
    \log(N+1)
    -
    \mathcal L_{\mathrm{DDAR}}.
\end{aligned}
\label{eq:chunk_nce_bound}
\end{equation}
Thus, under the idealized conditional negative-sampling assumption,
the action-wise alignment objective admits a lower-bound interpretation
for the conditional mutual information between Delta and the complete
action chunk.

\paragraph{Relation to Practical Negative Sampling.}
In practice, negative actions are sampled uniformly without replacement
from non-overlapping temporal regions of the same trajectory. This
sampling strategy keeps the task, scene, and embodiment context close
to that of the positive action while changing the execution stage. It
therefore prevents the contrastive task from being solved primarily
through task- or scene-level differences.

The practical negatives are not strictly independent samples from an
exact population marginal, so the lower bound in
Equation~\eqref{eq:chunk_nce_bound} should be interpreted under the
idealized sampling model above. The implemented loss is more precisely
an InfoNCE-based structured contrastive surrogate. Independently of the
mutual-information interpretation, however, the permutation invariance
in Equation~\eqref{eq:soft_delta_pooling}, the soft token assignments
in Equation~\eqref{eq:delta_action_assignment}, and the contrastive
gradient in Equation~\eqref{eq:nce_score_gradient} hold exactly for
the implemented objective. These properties encourage Delta to retain
transition information that distinguishes the current execution phase
from other actions within the same task trajectory.

\subsection{Limitations and Future Work}
\label{app:limitations}

While Vision-Language-Delta-Action (VLDA) provides a principled transition interface that improves manipulation performance, robustness, and long-horizon consistency, several limitations remain to be addressed in future work:

\paragraph{Sensitivity to Unrelated Visual Dynamics.}
The Local Transition Filter (LT-Filter) selects task-relevant changes based on the feature difference magnitude between current and future observations. Although the rich semantic priors from the V-JEPA 2 encoder significantly suppress pixel-level high-frequency noise, intense background motion (e.g., dynamic screens, moving bystanders) could occasionally pollute the top-$K$ selection. Future extensions could incorporate manipulation-centric visual grounding or instruction-attentive filters to explicitly isolate robot- and manipulation-centric regions from background clutter.

\paragraph{Fixed Horizon State Transition Modeling.}
Currently, VLDA models state transitions over a fixed temporal horizon $H$, dictated by the action chunking size. However, complex real-world tasks often involve variable-speed trajectories or multi-rate execution where state transitions naturally span non-uniform temporal spans. Extending the compact Delta interface to hierarchical or adaptively spaced temporal horizons represents a promising avenue for flexible long-horizon execution.

\paragraph{Cross-Embodiment Transition Transfer.}
Because the compact Delta representation focuses on physical world state changes rather than robot-specific joint configurations, it holds strong potential as an embodiment-agnostic control interface. Future work will investigate leveraging Delta targets for cross-robot skill transfer, where a transition model learned on one physical arm guides policy execution across diverse robotic platforms with minimal adaptation.



\end{document}